\documentclass[]{article}

\usepackage{proceed2e}
\usepackage{fancyhdr}
\usepackage{listings}
\usepackage{graphicx}
\usepackage{amsmath,amsbsy,amsthm,amssymb,amsfonts,bm}
\usepackage{epsfig}
\usepackage{epstopdf}
\usepackage{natbib}
\usepackage{caption}
\usepackage{subcaption}
\usepackage{tikz}
\usepackage{wasysym}
\usepackage{hyperref}
\usepackage{url}
\usetikzlibrary{arrows}

\newcommand{\vc}{\mathbf}

\title{Discriminative Switching Linear Dynamical Systems applied to Physiological Condition Monitoring}

\author{ {\bf Konstantinos Georgatzis} \\
School of Informatics \\
University of Edinburgh\\
\href{mailto:k.georgatzis@sms.ed.ac.uk}{\nolinkurl{k.georgatzis@sms.ed.ac.uk}} \\
\And
{\bf Christopher K. I. Williams} \\
School of Informatics \\
University of Edinburgh\\
\href{mailto:ckiw@inf.ed.ac.uk}{\nolinkurl{ckiw@inf.ed.ac.uk}} \\
}

\begin{document}

\maketitle

\begin{abstract} {
We present a Discriminative Switching Linear Dynamical System (DSLDS) applied to 
patient monitoring in Intensive Care Units (ICUs).
Our approach is based on identifying the state-of-health of a patient given
their observed vital signs using a discriminative classifier, and then
inferring their underlying physiological values conditioned on this status.
The work builds on the Factorial Switching Linear Dynamical System (FSLDS) \citep{quinn2009factorial}
which has been previously used in a similar setting. The FSLDS is a
generative model, whereas the DSLDS is a discriminative model. 
We demonstrate on two real-world datasets that the DSLDS is able to outperform the FSLDS in most cases of interest, and 
that an $\alpha$-mixture of the two models achieves higher performance than
either of the two models separately.}
\end{abstract}

Condition monitoring of patients in intensive care units (ICUs) based on vital signs (e.g.\ heart rate, blood pressure) is of critical importance, as they can be subject to a number of serious physiological events such as bradycardia and hypotension. However, a variety of artifactual processes can ``contaminate'' the data, e.g.\ the taking of blood samples, performing suctions, recalibrating sensors, etc. These artifactual processes complicate the task of identifying the important physiological events and are the main source of false alarms in ICUs. Moreover, it is of interest to maintain beliefs about the true physiological values of a patient when these cannot be directly observed due to artifact. For example, it would be desirable to display the patient's estimated blood pressure, when the corresponding measuring device has been disconnected or is otherwise displaying artifactual values (as is the case during a blood sample event). Of course, this estimate should be clearly distinguishable from the raw data (e.g. by using a different display colour).

One approach to this problem is to build a latent variable model, using a number of discrete latent variables to model the physiological and artifactual events through time, and a linear dynamical system (LDS) conditional on these discrete variables to model the associated dynamics in the vital signs observations. This is the factorial switching LDS (or FSLDS) of \citet{quinn2009factorial}. However, we have noticed that in building such systems it is necessary to construct quite detailed models of the artifactual events in order to capture them properly. This can be non-trivial since some of these events can be highly variable, which is hard to capture with a generative model. Despite this high variability, the vital signs can still contain informative features which could act as input to a discriminative model. Thus, if it is possible to build such a model that can fairly easily distinguish between the various events, then it would seem simpler and easier to make the discrete-state inference be discriminative, and use FSLDS-style inference for the continuous latent variables conditional on the inferred discrete state. We call this a \emph{discriminative switching linear dynamical system} (DSLDS). In this paper we compare the FSLDS and DSLDS models on two ICU condition monitoring datasets. The results show that using the DSLDS gives increased performance in most cases of interest, and that an $\alpha$-mixture of the two methods was able to achieve a higher performance than either of the two models separately.

To summarise, our goal is to build a model with increased performance for the following tasks:
 \begin{itemize}
  \item Identifying artifactual processes (e.g blood samples), which will reduce the high false alarm rate in ICUs and facilitate the task of identifying physiological processes.
  \item Identifying physiological processes which can be of critical importance (e.g bradycardias).
  \item Providing an estimate of a patient's true physiological values when these are obscured by artifact.
 \end{itemize}

The structure of the remainder of the paper is as follows: in Section \ref{sec:model} we give a description of our proposed model and compare its graphical structure and inference methods to those of the FSLDS, and briefly describe related work. In Section \ref{sec:experiments} we describe our experiments and provide results for the comparison between the DSLDS and the FSLDS. Finally, in Section \ref{sec:discussion} we conclude with general remarks about our proposed model and suggestions for future work. 

\section{Model description}
\label{sec:model}

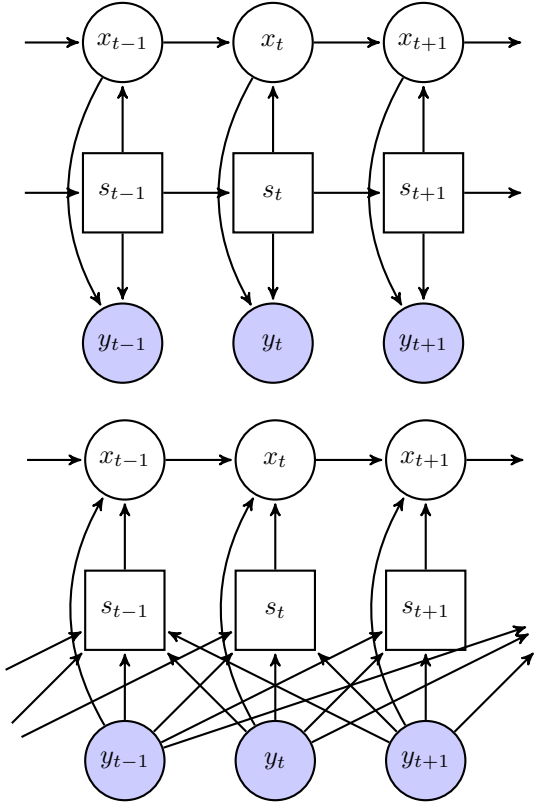
\begin{figure}[ht]
\hspace{-7mm}
\begin{subfigure}{.5\textwidth}
\begin{center}
\begin{tikzpicture}[->,>=stealth',shorten >=1pt,auto,node distance=2cm,
  thick,state node/.style={circle,fill=white!20,minimum size = 30pt,draw,font=\fontsize{10}{1.5}\selectfont},
  switch node/.style={rectangle,fill=white!20,minimum size = 30pt,draw,font=\fontsize{10}{1.5}\selectfont},
  obs node/.style={circle,fill=blue!20,minimum size = 30pt,draw,font=\fontsize{10}{1.5}\selectfont}]
  \node[switch node] (1) {$s_{t-1}$};
  \coordinate [left of =1] (h1);
  \node[switch node] (2) [right of=1] {$s_{t}$};
  \node[switch node] (3) [right of=2] {$s_{t+1}$};
  \coordinate [right of=3] (h3);
  \node[state node] (4) [above of=1] {$x_{t-1}$};
  \coordinate [left of =4] (h4);
  \node[state node] (5) [above of=2] {$x_{t}$};
  \node[state node] (6) [above of=3] {$x_{t+1}$};
  \coordinate [right of=6] (h6);
  \node[obs node] (7) [below of=1] {$y_{t-1}$};
  \node[obs node] (8) [below of=2] {$y_{t}$};
  \node[obs node] (9) [below of=3] {$y_{t+1}$};
  \draw [->,shorten <=0.7cm] (h1) to node {} (1);
  \draw [->,shorten <=0.7cm] (h4) to node {} (4);
  \draw [->,shorten >=0.7cm] (3)  to node {} (h3);
  \draw [->,shorten >=0.7cm] (6)  to node {} (h6);
  \path[every node/.style={font=\fontsize{1}{1.5}\selectfont}]
    (1) edge node [right] {} (2)
    (1) edge      [below] node {} (7)
    (1) edge node [above] {} (4)
    (2) edge node [right] {} (3)
    (2) edge node [above] {} (5)
    (2) edge      [below] node {} (8)
    (3) edge node [above] {} (6)
    (3) edge      [below] node {} (9)
    (4) edge node [right] {} (5)
    (4) edge [bend right] node [below] {} (7)
    (5) edge node [right] {} (6)
    (5) edge [bend right] node [below] {} (8)
    (6) edge [bend right] node [below] {} (9);
\end{tikzpicture}
\end{center}
\end{subfigure}%

\vspace{4mm}
\hspace{-24mm}
\begin{subfigure}{.5\textwidth}
\begin{center}
\begin{tikzpicture}[->,>=stealth',shorten >=1pt,auto,node distance=2cm,
  thick,state node/.style={circle,fill=white!20,minimum size = 30pt,draw,font=\fontsize{10}{1.5}\selectfont},
  switch node/.style={rectangle,fill=white!20,minimum size = 30pt,draw,font=\fontsize{10}{1.5}\selectfont},
  obs node/.style={circle,fill=blue!20,minimum size = 30pt,draw,font=\fontsize{10}{1.5}\selectfont}]
  \node[switch node] (1) {$s_{t-1}$};
  \node[switch node] (2) [right of=1] {$s_{t}$};
  \node[switch node] (3) [right of=2] {$s_{t+1}$};
  \coordinate [right of =3] (h3);
  \node[state node] (4) [above of=1] {$x_{t-1}$};
  \coordinate [left of =4] (h4);
  \node[state node] (5) [above of=2] {$x_{t}$};
  \node[state node] (6) [above of=3] {$x_{t+1}$};
  \coordinate [right of =6] (h6);
  \node[obs node] (7) [below of=1] {$y_{t-1}$};
  \coordinate [left of =7] (h7);
  \coordinate [left of =h7] (hh7);
  \node[obs node] (8) [below of=2] {$y_{t}$};
  \node[obs node] (9) [below of=3] {$y_{t+1}$};
  \draw [->,shorten <=0.7cm] (h4) to node {} (4);
  \draw [->,shorten <=0.7cm] (h7) to node {} (1);
  \draw [->,shorten <=2.7cm] (hh7) to node {} (1);
  \draw [->,shorten <=0.7cm] (h7) to node {} (2);
  \draw [->,shorten >=0.7cm] (6) to node {} (h6);
  \draw [->,shorten >=0.7cm] (7) to node {} (h3);
  \draw [->,shorten >=0.7cm] (8) to node {} (h3);
  \draw [->] (8) to node {} (1);
  \draw [->,shorten >=0.8cm] (9) to node {} (h3);
  \draw [->] (9) to node {} (1);
  \draw [->] (9) to node {} (2);
  \path[every node/.style={font=\fontsize{1}{1.5}\selectfont}]
    (7) edge node [below] {} (1)
    (7) edge [bend left] node {} (4)
    (1) edge node [below] {} (4)
    (2) edge node [below] {} (5)
    (8) edge [bend left] node {} (5)
    (3) edge node [below] {} (6)
    (9) edge [bend left] node {} (6)
    (4) edge node [right] {} (5)
    (7) edge node [below] {} (2)
    (7) edge node [below] {} (3)
    (5) edge node [right] {} (6)
    (8) edge node [below] {} (2)
    (8) edge node [below] {} (3)
    (9) edge node [below] {} (3);
\end{tikzpicture}
\end{center}
\end{subfigure}
\caption{Graphical model of the FSLDS (top) and the DSLDS (bottom). The state-of-health and underlying physiological values of a patient are represented by $s_t$ and $\vc{x}_t$ respectively. The shaded nodes correspond to the observed physiological values, $\vc{y}_t$. Note that in the case of the DSLDS the conditional probability $p(s_{t}|\vc{y}_{t-l:t+r})$ is modelled directly.}
\label{fig:DFSLDS}
\end{figure}

The graphical model of the FSLDS is depicted in Figure \ref{fig:DFSLDS} (top). It operates on three different sets of variables: The observed variables, $\vc{y}_t \in \mathbb{R}^{d_{y}}$ represent the patient's vital signs obtained from the monitoring devices at time $t$, which act as the input to our model. The continuous latent variables, $\vc{x}_t \in \mathbb{R}^{d_{x}}$, track the evolution of the dynamics of a patient's underlying physiology. The discrete variable, $s_t$, represents the switch setting or regime which the patient is currently in (e.g.\ stable, a blood sample is being taken etc.\ ). The switch variable can be factorised according to the cross-product of $M$ factors, so that $s_t = f_{t}^{1} \otimes f_{t}^{2} \otimes ... \otimes f_{t}^{M}$. Each factor variable, $f_{t}^{m}$, is usually a binary vector indicating the presence or absence of a factor, but in general it can take on $L^{(m)}$ different values and $K = \prod_{m=1}^M L^{(m)}$ is the total number of possible configurations of the switch variable, $s_t$. Also, $s_t$ depends explicitly on the previous time step, so that $p(s_t|s_{t-1}) = \prod_{m=1}^{M} p(f_{t}^{m}|f_{t-1}^{m})$. Conditioned on a particular regime, the FSLDS is equivalent to an LDS. The FSLDS can be seen then as a collection of LDS's, where each LDS models the dynamics of a patient's underlying physiology under a particular regime, and can also be used to generate a patient's observed vital signs. An LDS provides a generative framework for modelling our belief over the state space, given observations. 

We can alternatively adopt a discriminative view. We start by modelling $p(s_t|\vc{y}_{t-l:t+r})$ with a discriminative classifier, where (features of) observations from the previous $l$ and future $r$ time steps affect the belief of the model about $s_t$. The inclusion of $r$ frames of future context is analogous to fixed-lag smoothing in an FSLDS (see e.g.\ \citealp[sec.\ 10.5]{sarkka2013bayesian}). We note that inclusion of future observations in the conditioning set means that the DSLDS will operate with a delay of $r$ seconds, since an output of the model at time $t$ can be produced only after time $t+r$. Provided that $r$ is small enough ($r\leq$10 in experiments), this delay is negligible compared to the increase in performance. The LDS can also be regarded from a similarly discriminative viewpoint which allows us to model $p(\vc{x}_t|\vc{x}_{t-1},\vc{y}_t)$. This is similar to the Maximum Entropy Markov Model (MEMM) \citep{mccallum2000maximum} with the difference that the latent variable is continuous rather than discrete. The main advantage of this discriminative view is that it allows for a rich number of (potentially highly correlated) features to be used without having to explicitly model their distribution or the interactions between them, as is the case in a generative model. A combination of these two discriminative viewpoints gives rise to the DSLDS graphical model in Figure \ref{fig:DFSLDS} (bottom). The DSLDS, conditioned on $s_t$, can be seen then as a collection of MEMM's, where each MEMM in the DSLDS plays a role equivalent to that of each LDS in the FSLDS.

The DSLDS can be defined as 
\begin{align}
\label{eq:model}
\nonumber p(\vc{s},\vc{x}|\vc{y}) = \hspace{1mm} &p(s_1|\vc{y}_1)p(\vc{x}_1|s_1,\vc{y}_1) \times \\
                                  \hspace{-2mm}&\prod_{t=2}^T p(s_t|\vc{y}_{t-l:t+r})p(\vc{x}_t|\vc{x}_{t-1},s_t,\vc{y}_t) \hspace{2mm}.
\end{align}

The simplest assumption we can make for the DSLDS is that $p(s_t|\vc{y}_{t-l:t+r})$ factorises, so that 
 
\begin{align}
\label{eq:factorised_switch}
 p(s_t|\vc{y}_{t-l:t+r}) = \prod_{m=1}^M p({f}_{t}^{(m)}|\vc{y}_{t-l:t+r}) \hspace{2mm} .  
\end{align}

However, one could use a structured output model to predict the joint distribution of different factors.

\subsection{Predicting s$_{t}$}
\label{sec:s_t}

Our belief about the state of health of a patient at time $t$ is modelled by $p(s_t|\vc{y}_{t-l:t+r})$, the conditional probability of the switch variable given the observed vital signs. Following the factorisation of the switch variable in eq.\ \ref{eq:factorised_switch}, we model the conditional probability of each factor being active at time $t$ given the observations with a probabilistic discriminative binary classifier, so that $p({f}_{t}^{(i)} = 1|\vc{y}_{t-l:t+r}) = G(\phi(\vc{y}_{t-l:t+r}))$, where $G(\cdot)$ is a classifier-specific function, and $\phi(\vc{y}_{t-l:t+r})$ is the feature vector that acts as input to our model at each time step as described in Section \ref{sec:features}. As is evident from Figure \ref{fig:DFSLDS} (bottom) there is no explicit temporal dependence on the switch variable sequence. However, temporal continuity is implicitly incorporated in our model through the construction of the features.

\subsubsection{An $\alpha$-mixture of s$_{t}$}
The DSLDS model can be seen as complementary to the FSLDS, and they can be run in parallel. One way of combining the two outputs is to maintain an $\alpha$-mixture over $s_t$. If $p_{g}(s_t)$ and $p_{d}(s_t)$ are the outputs for the switch variable at time $t$ from FSLDS and the DSLDS respectively, then their $\alpha$-mixture is given by: $p_{\alpha}(s_t) = c \left( {p_{g}(s_t) }^{(1-\alpha)/2} + {p_{d}(s_t)}^{(1-\alpha)/2} \right)^{2/(1-\alpha)}$, where $c$ is a normalisation constant which ensures that $p_{\alpha}(s_t)$ is a probability distribution. The family of $\alpha$-mixtures then subsumes various known mixtures of distributions and defines a continuum across them via the $\alpha$ parameter. For example, for $\alpha=-1$ we retrieve the mixture of experts (with equally weighted experts) framework, while for $\alpha \rightarrow 1$, the formula yields $p_{1}(s_t) = c\sqrt{p_{g}(s_t)p_{d}(s_t)}$, rendering it equivalent to a product of 
experts viewpoint. In general, as $\alpha$ increases, the $\alpha$-mixture assigns more weight to the smaller elements of the mixture (with $\alpha\rightarrow\infty$ giving $p_{\infty}(s_t) = \min\{p_{g}(s_t),p_{d}(s_t)\}$), while as $\alpha$ decreases, more weight is assigned to the larger elements (with $\alpha\rightarrow - \infty$ giving $p_{-\infty i}(s_t) = \max\{p_{g}(s_t),p_{d}(s_t)\}$) A thorough treatment is given in \citet{amari2007integration}.

\subsection{Predicting x$_{t}$}
\label{sec:x_t}
The model of the patient's physiology should capture the underlying temporal dynamics of their observed vital signs under their current health state. The idea is that the current latent continuous state of a patient should be dependent on (a) the latent continuous state at the previous time step, (b) the current state of health and (c) the current observed values. We model these assumptions as follows

\begin{align}
\label{eq:state_space}
 \nonumber &p(\vc{x}_t|\vc{x}_{t-1},s_t,\vc{y}_t) \propto \\ 
 \nonumber &\exp\lbrace \scalebox{0.75}[1.0]{\( - \)}\dfrac{1}{2}(\vc{x}_t \scalebox{0.75}[1.0]{\( - \)} \vc{A}^{(s_{t})}\vc{x}_{t-1})^{\top} (\vc{Q}^{(s_{t})})^{-1} (\vc{x}_t \scalebox{0.75}[1.0]{\( - \)} \vc{A}^{(s_{t})}\vc{x}_{t-1})\rbrace \times\\
 &\exp \lbrace \scalebox{0.75}[1.0]{\( - \)}\dfrac{1}{2}(\vc{C}^{(s_{t})}\vc{x}_t \scalebox{0.75}[1.0]{\( - \)} \vc{y}_t)^{\top} (\vc{R}^{(s_{t})})^{-1} (\vc{C}^{(s_{t})}\vc{x}_{t} \scalebox{0.75}[1.0]{\( - \)} \vc{y}_t)\rbrace \hspace{2mm}.
\end{align}

The first term on the RHS of eq.\ \ref{eq:state_space} is the {\bf system model} for an LDS and captures the dynamics of a patient's latent physiology under state $s_{t}$. The second term can be seen as the discriminative counterpart of the {\bf observation model} of an LDS. In our condition monitoring setting, the observed vital signs are considered to be noisy realisations of the true, latent physiology of a patient and thus, the observation model encodes our belief that $\vc{x}_{t}$ is a noisy version of $\vc{y}_{t}$. Under this assumption, $\vc{C}^{s_t}$ consists of 0/1 entries, which are set based on our knowledge of whether the observations $\vc{y}_{t}$ are artifactual or not under state $s_t$. In the FSLDS, the corresponding observation model encodes the belief that the generated $\vc{y}_{t}$ should be normally distributed around $\vc{x}_{t}$ with covariance $\vc{R}^{s_t}$, whereas in our discriminative version, the observation model encodes our belief that $\vc{x}_{t}$ should be normally distributed around $\vc{y}_{t}$ with covariance $\vc{R}^{s_t}$. The idea behind this model is that at each time step we update our belief about $\vc{x}_t$ conditioned on its previous value, $\vc{x}_{t-1}$, and the current observation, $\vc{y}_t$, under the current regime $s_t$. For example, under an artifactual process, the observed signals do not convey useful information about the underlying physiology of a patient. In that case, we drop the connection between $\vc{y}_t$ and $\vc{x}_t$ (for the artifact-affected channels) which translates into setting the respective entries of $\vc{C}^{s_t}$ to zero. Then, the latent state $\vc{x}_t$ evolves only under the influence of the appropriate system dynamics parameters $(\vc{A}^{(s_{t})},\vc{Q}^{(s_{t})})$. Conversely, operation under a non-artifactual regime incorporates the information from the observed signals, effectively transforming the inferential process for $\vc{x}_t$ into a product of two ``experts'', one propagating probabilities from $\vc{x}_{t-1}$ and one from the current observations. 

We note that the step of conditioning on the current regime $s_t$ in order to predict $\vc{x}_t$ is required for our task, as we do not have training data for the $\vc{x}$-state. Otherwise, one could imagine building a simpler model such as a conditional random field \citep{lafferty2001conditional}, to predict the $\vc{x}$-state directly from the observations. However, in our case, where only labels about the patient's regime are available, this is not possible.

\subsection{Learning}
\label{sec:learning}
We first describe learning in the general SLDS setting. The parameters that need to be learned are: \{$\vc{A}^{s}$, $\vc{Q}^{s}$, $\vc{C}^{s}$, $\vc{R}^{s}$\}. Given training data for each switch setting, these can be learned independently as LDS parameters for each configuration of $s$. Following \citet{quinn2009factorial} we use an independent ARIMA model with added observation noise for each channel. Casting such a model into state space form is a standard procedure as described in \citet[sec.\ 12.1]{brockwell2009time}, and amounts into reformulating the parameters of the ARIMA model into the parameters of a state-space model. Once the model is in state space form, $\vc{A}^{s}$, $\vc{Q}^{s}$, $\vc{C}^{s}$, $\vc{R}^{s}$ can be fit according to the maximum likelihood criterion by using numerical optimisation methods (like Newton-Raphson, Gauss-Newton), as presented in \citet[sec.\ 2.6]{shumway2000time} or expectation maximisation (EM) as presented in \citet{ghahramani1996parameter}. We note that the vector ARMA (VARMA) representation is used, where for example a one-dimensional AR($p$) process can be encoded as a $p+1$-dimensional VAR(1) process by maintaining a latent state representation of the form $\vc{x}_{t} = [{x}_{t}\;{x}_{t-1}\;...\;{x}_{t-p}]$. 

In the DSLDS, the same set of parameters needs to be learned. As mentioned in Section \ref{sec:x_t}, the assumptions for the DSLDS observation model constrain $\vc{C}^{s}$ to be a binary matrix, whose values are set so as to pick the most recent value $\vc{x}_{t}$ under the VARMA representation. For example, assuming that we are modelling one channel, under a physiological regime, as an AR(2) process, then $\vc{C}^{s} = [1\;0\;0]$. Under this constrained form of $\vc{C}^{s}$ we obtain the remaining parameters, $\vc{A}^{s}$, $\vc{Q}^{s}$ and $\vc{R}^{s}$, using the same learning process as the one already described for the case of a general SLDS. 

The task of determining the order of the respective ARIMA models is less straightforward. We have followed a practical approach as suggested in \citet[sec.\ 6.2]{diggle1990time}. The autocorrelation and partial autocorrelation function (ACF and PACF respectively) of the stationary data (if a time series is not stationary, we make it stationary by successive differencing) were examined to provide an initial estimate of the appropriate model order. A clear cut-off at lag $q$ in the ACF plot is suggestive of an MA($q$) process, while a clear cut-off at lag $p$ in the PACF plot is suggestive of an AR($p$) process. Clear cut-offs are rare in a real world application, in which case we looked for less clear tail-offs in the PACF and ACF plots. After establishing a small number of potential model orders suggested by these tail-offs, further exploration of the model order around these initial estimates was carried out by calculating the Akaike Information Criterion (AIC) score \citep{akaike1972information} for each of these potential model orders, and finally the one with the smallest AIC value was chosen.

\subsection{Inference}
\label{sec:inference}
In this paper we are concerned with the task of computing the distribution $p(s_t,\vc{x}_t|\vc{y}_{1:t+r})$. According to our proposed model, $p(s_t|\vc{y}_{t-l:t+r})$ can be inferred at each time step via a classifier as described in Section \ref{sec:s_t}. However, exact inference for $\vc{x}_t$ is still intractable. The same limitation as in the case of a standard SLDS applies \citep{lerner2001inference}: In order to maintain an exact belief over the posterior distribution of $\vc{x}_t$ we need to keep track of all the potential combinations of switch variable settings that could have lead us from $\vc{x}_{t-1}$ to $\vc{x}_t$, making inference scale exponentially with time. An approximation of this distribution can be maintained via the Gaussian Sum algorithm\footnote{The Gaussian Sum algorithm is also known as the Generalised Pseudo Bayesian (GPB) algorithm as mentioned in \citet{murphy1998switching}.} \citep{alspach1972nonlinear}. The idea is that at each time step $t$ we maintain an approximation of $p(\vc{x}_t|s_t,\vc{y}_{1:t+r})$ as a mixture of $J$ Gaussians. Moving one time step forward will result in the posterior $p(\vc{x}_{t+1}|s_{t+1},\vc{y}_{1:t+r+1})$ having $KJ$ components, which are again collapsed to $J$ components. In our experiments we use $J=1$, which translates into matching moments (up to second order) of the distribution for each setting of $s_t$, as shown in \citet{murphy1998switching}. Therefore inference in the DSLDS can be seen as a two-step process, where $p(s_t|\vc{y}_{t-l:t+r})$ is inferred by our discriminative classifier, and $p(\vc{x}_t|s_t,\vc{y}_{1:t+r})$ is inferred according to the Gaussian Sum algorithm.

\subsection{Related work}
\label{sec:review}
In terms of methodology, our proposed model bears some similarities to the one used by \citet{lu2009hybrid}. However, their model was used to model spatial relationships and they were only concerned with a binary discrete latent space. In our case, we are concerned with modelling temporal structure and we have a richer and more complex discrete latent space. More importantly, in their work the distribution maintained over the continuous latent space is a single multivariate Gaussian, whereas in our model, as described in the previous section, the belief over the continuous latent space is modelled as a mixture of $KJ$ Gaussians. This allows us to keep track of multiple modes about the belief over a patient's underlying physiology, since this is potentially affected by multiple factors. 

In terms of application, our work is mostly similar to the one presented in \citet{quinn2009factorial}. The same task of inferring artifactual and physiological processes was considered there. However a generative approach was taken there via the use of an FSLDS. In our case, we take a discriminative approach, which performs better in the experiments considered below. Also, in \citet{lehman2014physiological}, a switching vector autoregressive model was used on minute-by-minute heart rate and blood pressure vital signs to provide inputs for a logistic regression classifier with the goal of patient outcome prediction. In our work, we use a more expressive model, capable of modelling both discrete and continuous latent states under a unified framework, for the purposes of detecting patients' state-of-health and inferring their underlying physiology.  

\section{Experiments}
\label{sec:experiments}
In this section we describe experiments on two challenging datasets comprising of patients admitted to ICUs in two different hospitals, namely a neonatal ICU and an adult ICU. We emphasise that it is highly non-trivial to obtain annotations for medical datasets as it requires the very scarce resource of experienced clinicians. Indeed, for the adult ICU, the annotated data are the product of a one-year collaboration with that ICU. Physionet \citep{goldberger2000physiobank}, a freely available medical dataset, is not suitable for our task since the only available time-series annotations are a limited set of life threatening/terminal events, for which identification would not be of practical use in the ICU. 

For both datasets, we evaluate the performance of the DSLDS compared to the FSLDS. We also report the performance of an $\alpha$-mixture of the two models. Note that the FSLDS has been shown in \citet{quinn2009factorial} to achieve superior results compared to more basic models such as a factorial hidden Markov model (FHMM) for the task of condition monitoring in ICUs. We first provide a short description of the various features that were used as input to the state-of-health model as described in Section \ref{sec:s_t}, followed by an outline of the main characteristics of the two datasets. We conclude this section by providing results on two tasks: a) inferring a patient's state of health and b) inferring a patient's underlying physiology in the presence of artifact corruption.   

\subsection{Features \& Classifiers}
\label{sec:features}
As described in Section \ref{sec:s_t}, the estimate of $s_t$ is the output of a discriminative classifier. For both datasets, we found that using a random forest \citep{breiman2001random} as our classification method yields the best performance. Suggestions for judicious selection of various tree-construction parameters can be found in \citet[Ch.\ 15]{hastie2009elements}. The Gini index was used as the criterion for splitting nodes for each tree in the random forest. The output of the random forest for a new test point is an average of the predictions produced by each tree, where the prediction of each tree is the proportion of the observations that belong to the positive class in the leaf node in which the test point belongs to. Apart from their high performance, another appealing property of random forests is that they can handle missing observations via the construction of surrogate variables and splits within each decision tree as explained in \citet[sec.\ 9.2.4]{hastie2009elements}.

We use a variety of features to capture interesting temporal structure between successive observations. At each time step, a sliding window of length $l+r+1$ is computed. For some features we also divide the window into further sub-windows and extract additional features from them. More precisely, the full set of features that are being used are: (i) the observed, raw values of the previous $l$ and future $r$ time steps ($\vc{y}_{t-l:t+r}$); (ii) the slopes (calculated by least squares fitting) of segments of that sliding window that are obtained by dividing it in segments of length $(l+r+1)/k$; (iii) an exponentially weighted moving average of this window of raw values (with a kernel of width smaller than $l+r+1$); (iv) the minimum, median and maximum of the same segments; (v) the first order differences of the original window; and (vi) differences of the raw values between different channels.

\subsection{Neonatal ICU}
\label{sec:NICU_data}
The first dataset is the one used in \citet{quinn2009factorial}\footnote{The dataset has been anonymised and is available at: \emph{www.cit.mak.ac.ug/staff/jquinn/software.html}}. It comprises 24-hour periods from fifteen neonates admitted to the ICU of the Edinburgh Royal Infirmary, with events of interest annotated by two clinical experts. These annotations include: i) blood sample events (BS), ii) periods during which an incubator is open (IO), iii) core temperature probe disconnections (TD), iv) bradycardias (BR), and v) periods that are clearly not stable but no further identification was made by the clinicians (X). These last cases can be collectively considered as a ``none-of-the-above'' factor, which is referred to as the X-factor by \citet{quinn2009factorial}. More details about the events of interest can be found in the aforementioned work. We used the same parameters for the underlying physiology model as the ones used there. 

\subsection{Adult ICU}
\label{sec:SGH_data}
The second dataset comprises data collected from nine adults admitted to the neuro ICU of the Southern General Hospital in Glasgow. An average of 33-hour periods were collected from each of these patients, consisting of measurements recorded on a second-by-second basis for four different channels: heart rate (HR), systolic and diastolic blood pressure (BPsys, BPdia), and systolic intracranial pressure (ICPsys). These data were then annotated by a clinical expert. We give a brief description of the learning process for stability periods and modelled factors, which include blood samples, damped traces (DT), suction events (SC), and the X-factor.

{\bf Stable periods} correspond to time periods when no annotation occurred from the experts, suggesting that the patient is in a stable condition. In \citet{williams2011automating} it was found that in a similar setting a 15 minute period of stability provides an adequate amount of training data. We use the same time interval for our experiments. We found that ARIMA(2,1,0) models were adequate for all channels.

An example of a {\bf blood sample} is shown in Figure \ref{fig:SGH_damped_BS} (bottom). Changes in BPsys and BPdia can be modelled as a four-stage process: i) the blood is diverted to a syringe for blood sampling, which causes an artifactual ramp in the observed measurements. This is similar to the blood sample model described in \citet{quinn2009factorial} and we follow the same approach here. ii) A recalibration stage follows, causing measurements to drop to zero which can be modelled similarly to a dropout event as in \citet{quinn2009factorial}. iii) BP measurements continue as a stable period for a brief period. iv) The blood sample is concluded with a flushing event for hygiene purposes which causes a sharp increase in measurements. This stage is modelled as an AR(3) process for both the BPsys and BPdia channels. A total number of 64 blood sample events have been annotated, with an average duration of 1.6 minutes.

During a {\bf suction event}, a flexible catheter is inserted into the airway of the patient to remove secretions that have accumulated over time in their pulmonary system. This event is observed as a significant increase in the values of all observed channels. An AR(2) process models the HR channel, while AR(3) processes were used to model the remaining channels. A total number of 53 suction events have been annotated, with an average duration of 4.3 minutes.

A {\bf damped trace}, an example of which is shown in Figure \ref{fig:SGH_damped_BS} (top), is usually observed due to blood residues being accumulated in the line used for measuring the blood pressure channel, which leads both BPsys and BPdia to converge to a similar mean value while at the same time the measurements exhibit high variability. Both channels were modelled with AR(3) processes. A total number of 32 damped trace events have been annotated, with an average duration of 14 minutes.

Except for the aforementioned factors which we explicitly model, there are a multitude of other factors present in our training data, corresponding to either known but not yet modelled factors (such as hygiene events, tachycardias etc.) or to unknown factors (clear abnormalities which however have not been identified by the clinicians). We collectively treat those events as unknown and model them according to the X-factor model proposed in \citet{quinn2009factorial}. A total number of 278 X-factor events have been annotated, with an average duration of 7.5 minutes. Channels which are unaffected by an artifactual process (as shown in Table \ref{tab:channels}) are modelled as in the stable case. In every case the parameters of the $\vc{x}$-state models were further optimised by EM.

\renewcommand{\tabcolsep}{0.11cm}
\begin{table}[ht]
\caption{Channels affected by different processes for the adult ICU are marked by $\bullet$.}
\label{tab:channels}
\begin{center}
\begin{tabular}{c l l l l}
\multicolumn{1}{c}{}  &\multicolumn{1}{c}{HR}  &\multicolumn{1}{c}{BPsys} &\multicolumn{1}{c}{BPdia} &\multicolumn{1}{c}{ICPsys}
\\ \hline \\
Blood sample          &${}$            &$\bullet$          &$\bullet$         &${}$                      \\
Damped trace          &${}$            &$\bullet$          &$\bullet$         &${}$                      \\
Suction               &$\bullet$          &$\bullet$          &$\bullet$         &$\bullet$                      \\
X-factor              &$\bullet$          &$\bullet$          &$\bullet$         &$\bullet$                      \\
\end{tabular}
\end{center}
\end{table}

\renewcommand{\tabcolsep}{0.11cm}
\begin{table}[ht]
\caption{Comparison of DSLDS, FSLDS and $\alpha$-mixture performance for the Neonatal ICU dataset. Optimal value of the $\alpha$ parameter is shown inside parenthesis.}
\label{tab:NICU}
\begin{center}
\begin{tabular}{c l l l l l}
\multicolumn{1}{c}{\bf AUC}  &\multicolumn{1}{c}{BS} \hspace{-1mm} &\multicolumn{1}{c}{IO} &\multicolumn{1}{c}{TD} &\multicolumn{1}{c}{BR} &\multicolumn{1}{c}{X}
\\ \hline \\
DSLDS                      \hspace{2mm}       &$0.98$   \hspace{1mm}        &$0.83$  \hspace{1mm}         &$0.90$ \hspace{1mm}         &$0.94$  \hspace{1mm}        &$0.57$           \\
FSLDS                      \hspace{2mm}       &$0.92$           &$0.87$           &$0.88$          &$0.85$          &$0.66$           \\
$\alpha$-mixture$^{(0.5)}$ \hspace{2mm}       &$0.98$           &$0.89$           &$0.93$          &$0.92$          &$0.67$  \\
\end{tabular}
\end{center}
\end{table}

\subsection{Results}

\begin{figure}[ht]
        \begin{center}
        \begin{subfigure}[b]{0.25\textwidth}
                \centering
                \includegraphics[width=\textwidth, height = 6.5cm, keepaspectratio=true]{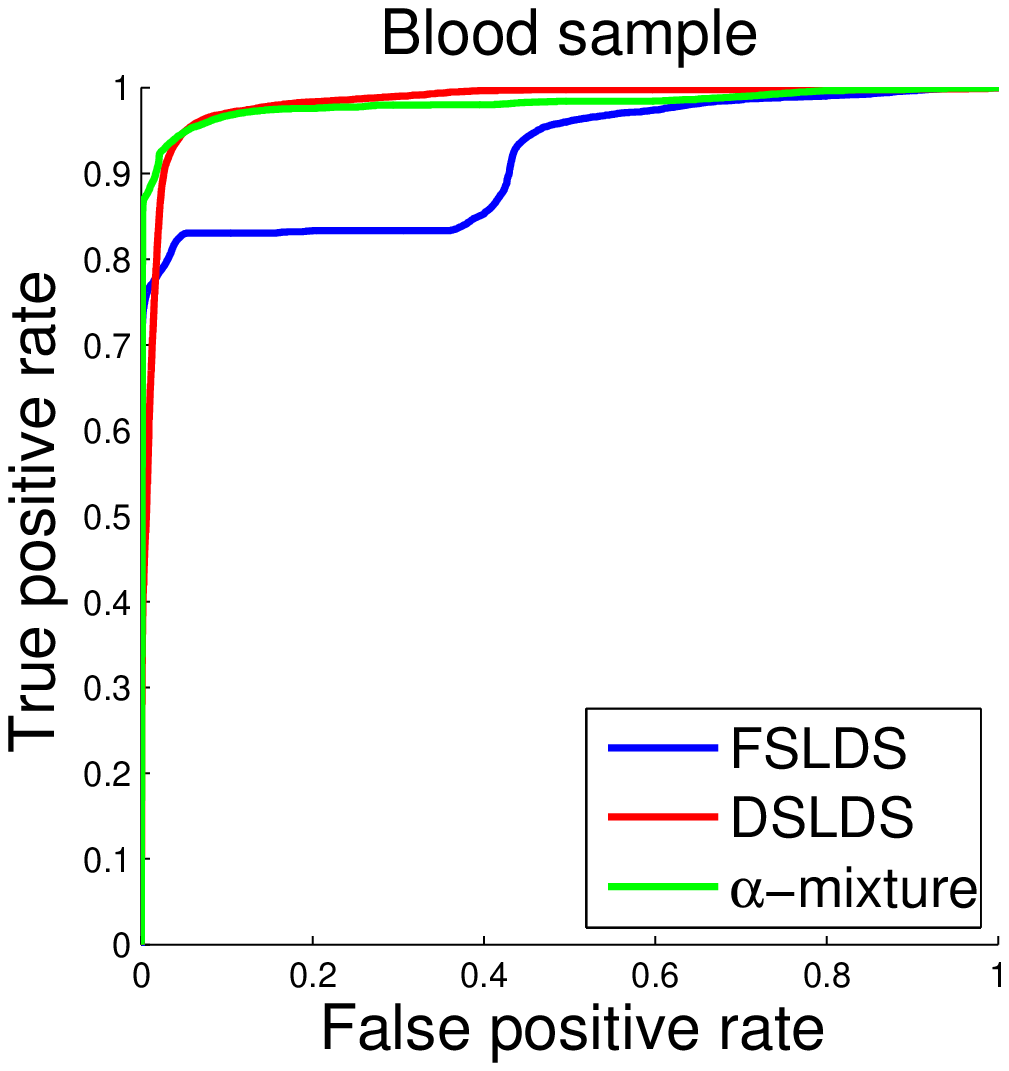}
        \end{subfigure}%
        ~         
        \begin{subfigure}[b]{0.25\textwidth}
                \centering
                \includegraphics[width=\textwidth, height = 6.5cm, keepaspectratio=true]{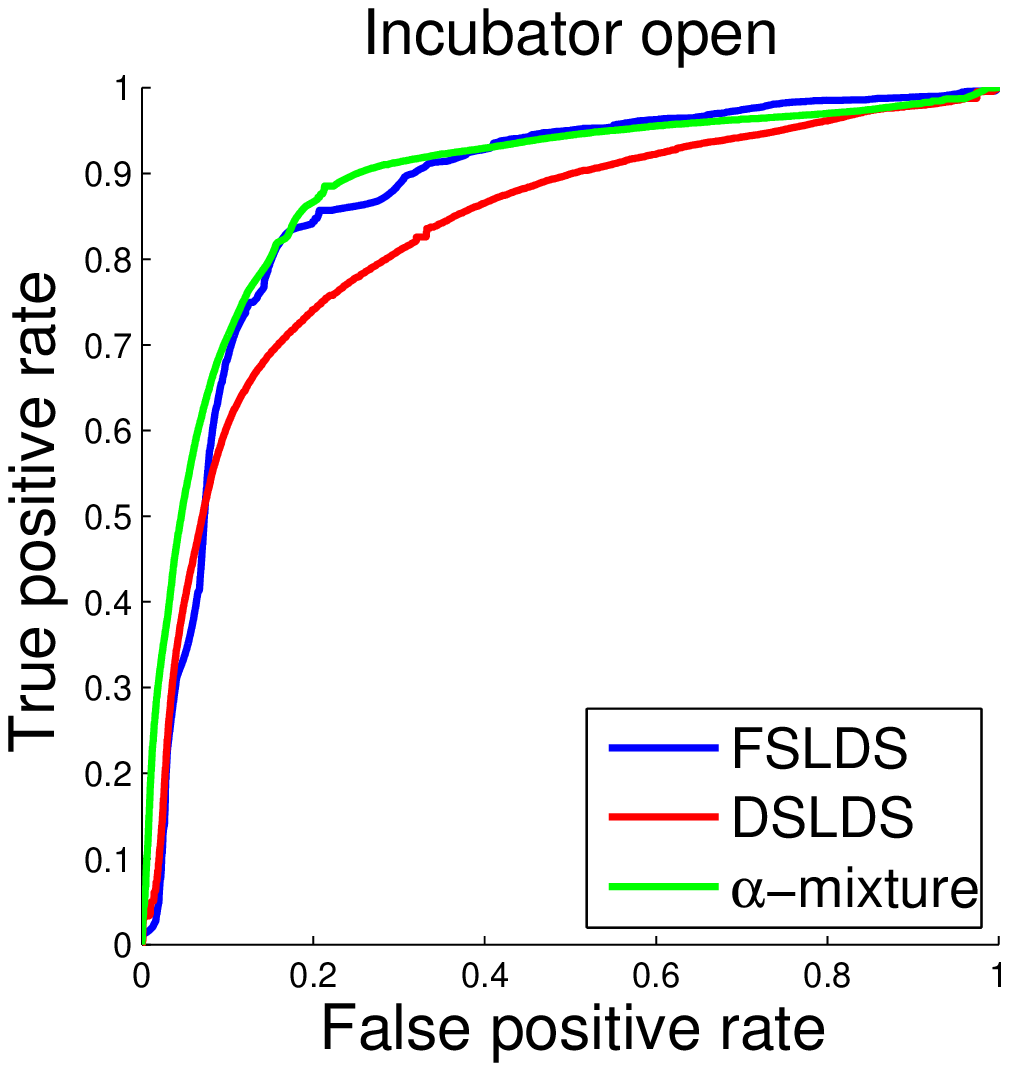}
        \end{subfigure}
        
        \begin{subfigure}[b]{0.25\textwidth}
                \centering
                \includegraphics[width=\textwidth, height = 6.5cm, keepaspectratio=true]{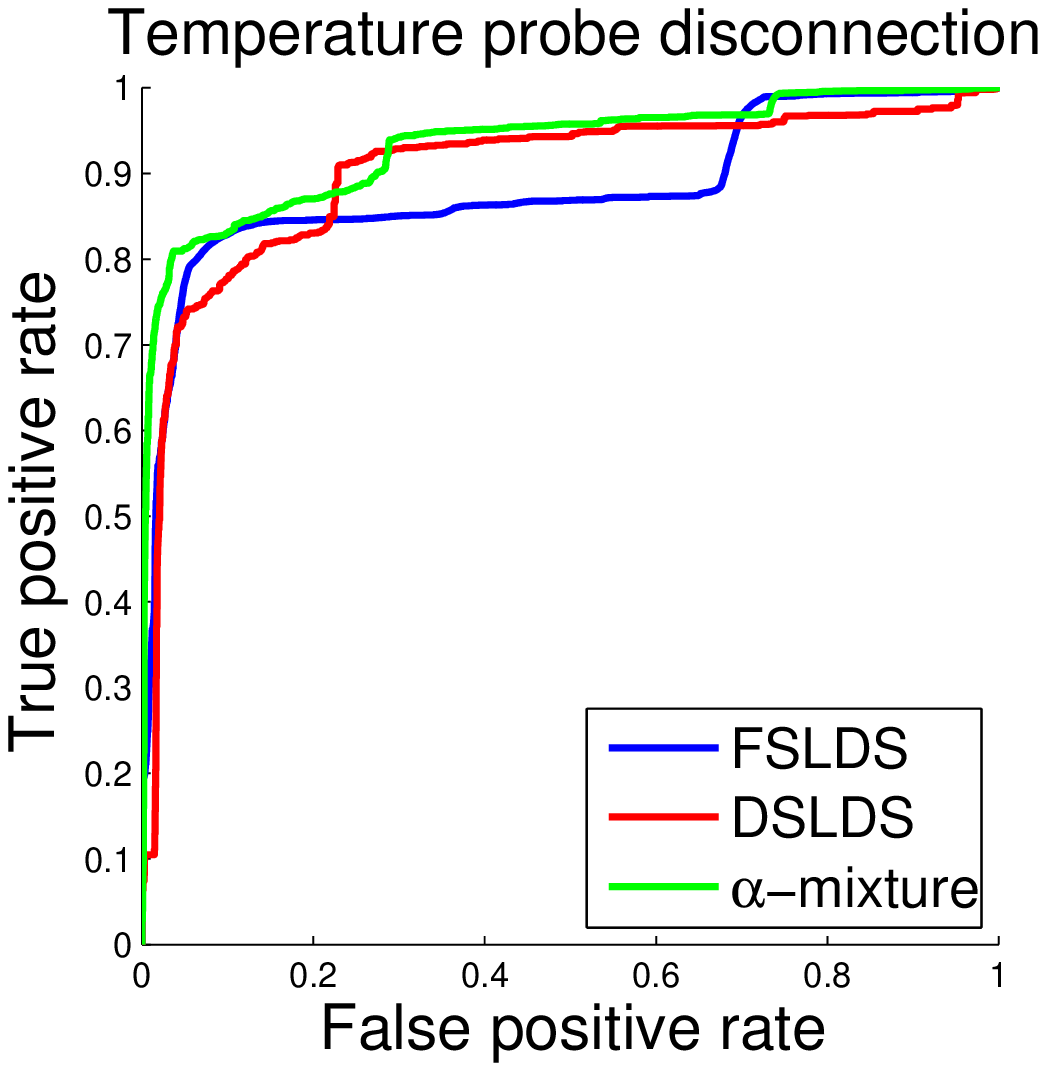}
        \end{subfigure}%
        ~         
        \begin{subfigure}[b]{0.25\textwidth}
                \centering
                \includegraphics[width=\textwidth, height = 6.5cm, keepaspectratio=true]{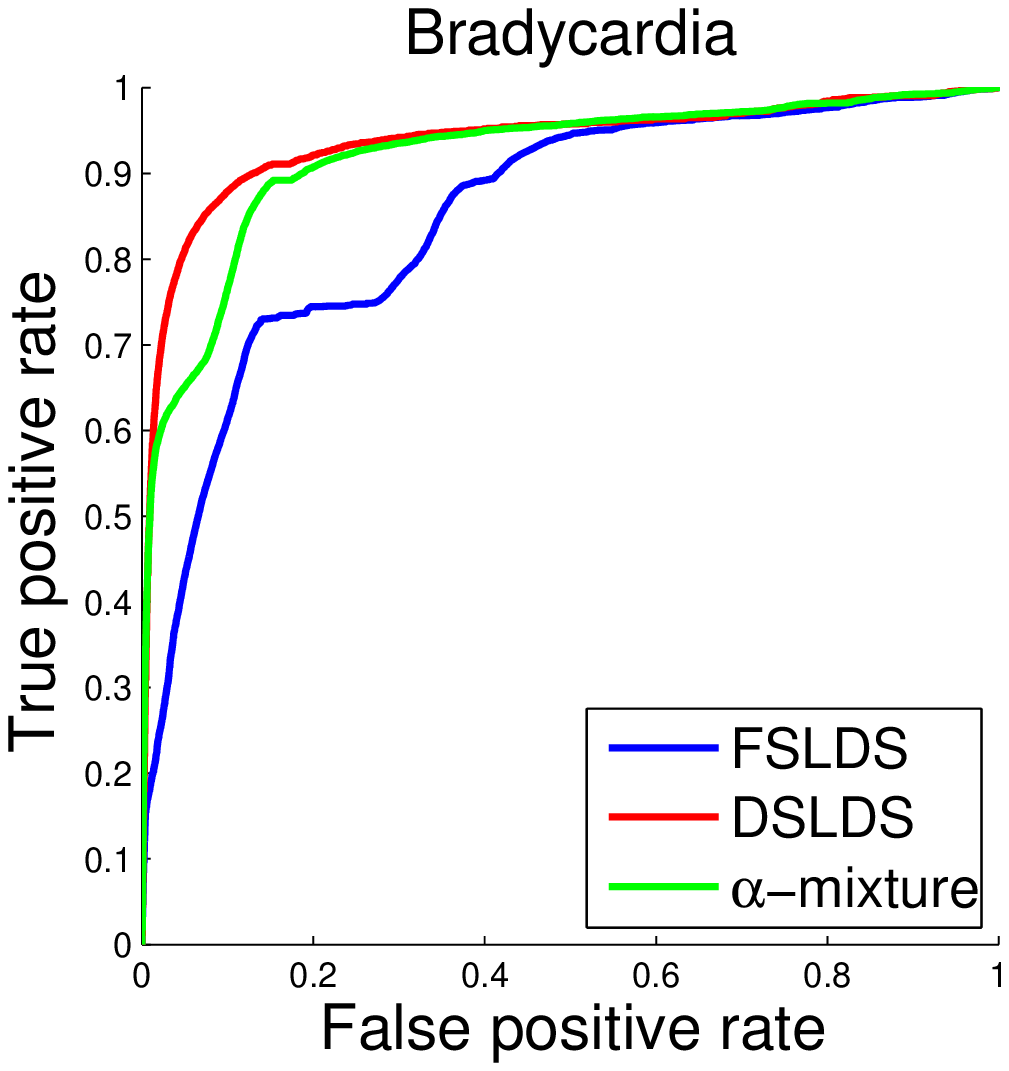}
        \end{subfigure}

        \begin{subfigure}[b]{0.25\textwidth}
                \centering
                \includegraphics[width=\textwidth, height = 6.5cm, keepaspectratio=true]{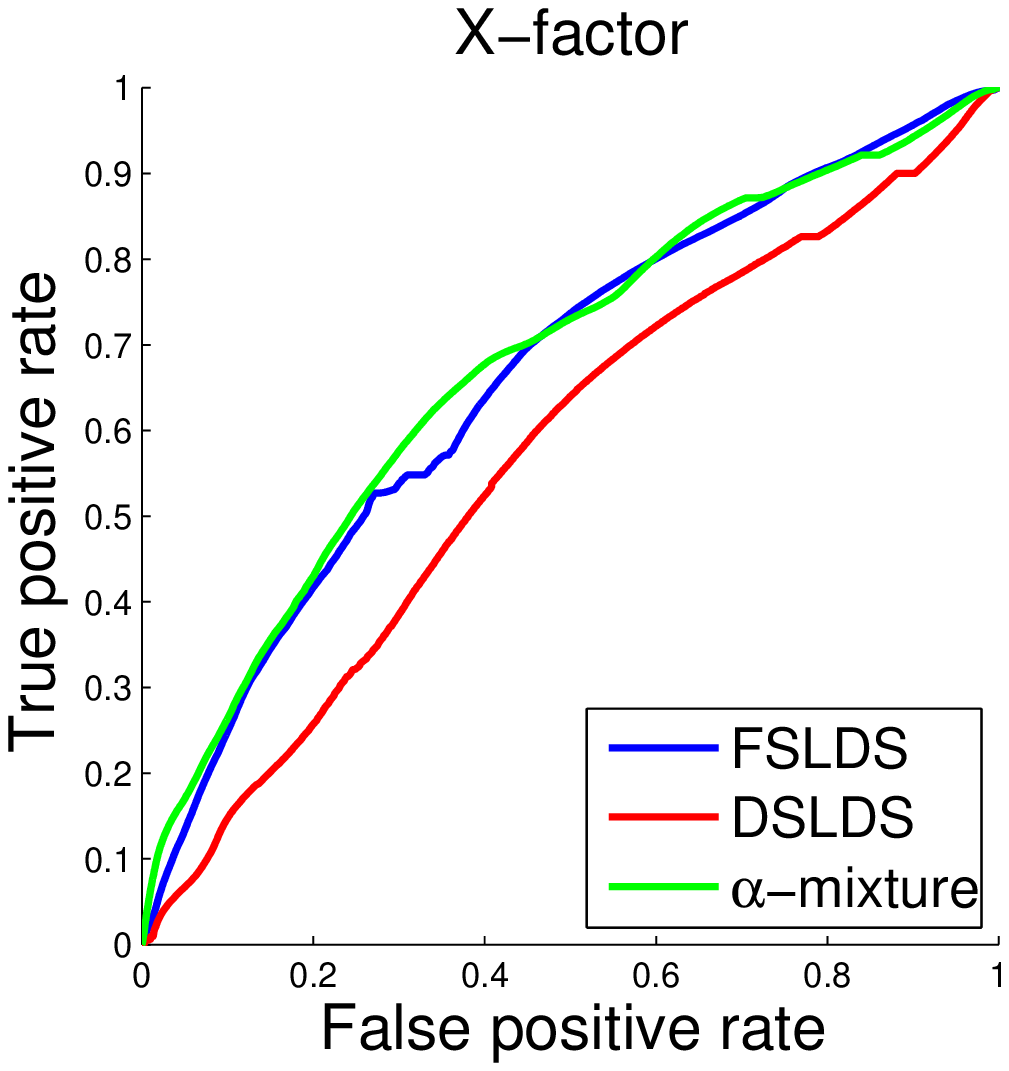}
        \end{subfigure}%
        ~         
        \begin{subfigure}[b]{0.25\textwidth}
                \centering
        \end{subfigure}
        \caption{ROC curves per modelled factor in the case of the neonatal ICU.}
        \label{fig:ROC_nicu}
        \end{center}
\end{figure}

\begin{figure}[hm]
        \begin{center}
        \begin{subfigure}[b]{0.25\textwidth}
                \centering
                \includegraphics[width=\textwidth, height = 6.5cm, keepaspectratio=true]{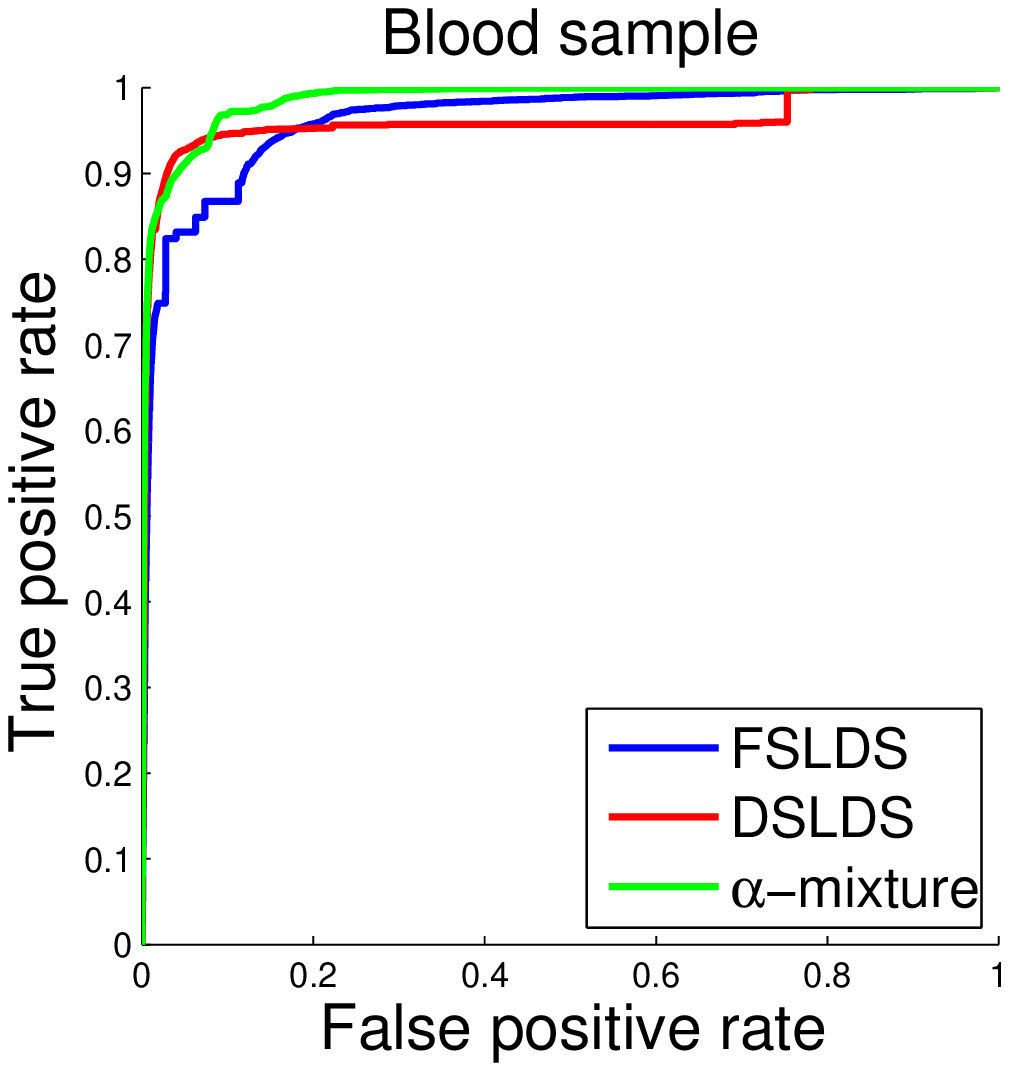}
        \end{subfigure}%
        ~        
        \begin{subfigure}[b]{0.25\textwidth}
                \centering
                \includegraphics[width=\textwidth, height = 6.5cm, keepaspectratio=true]{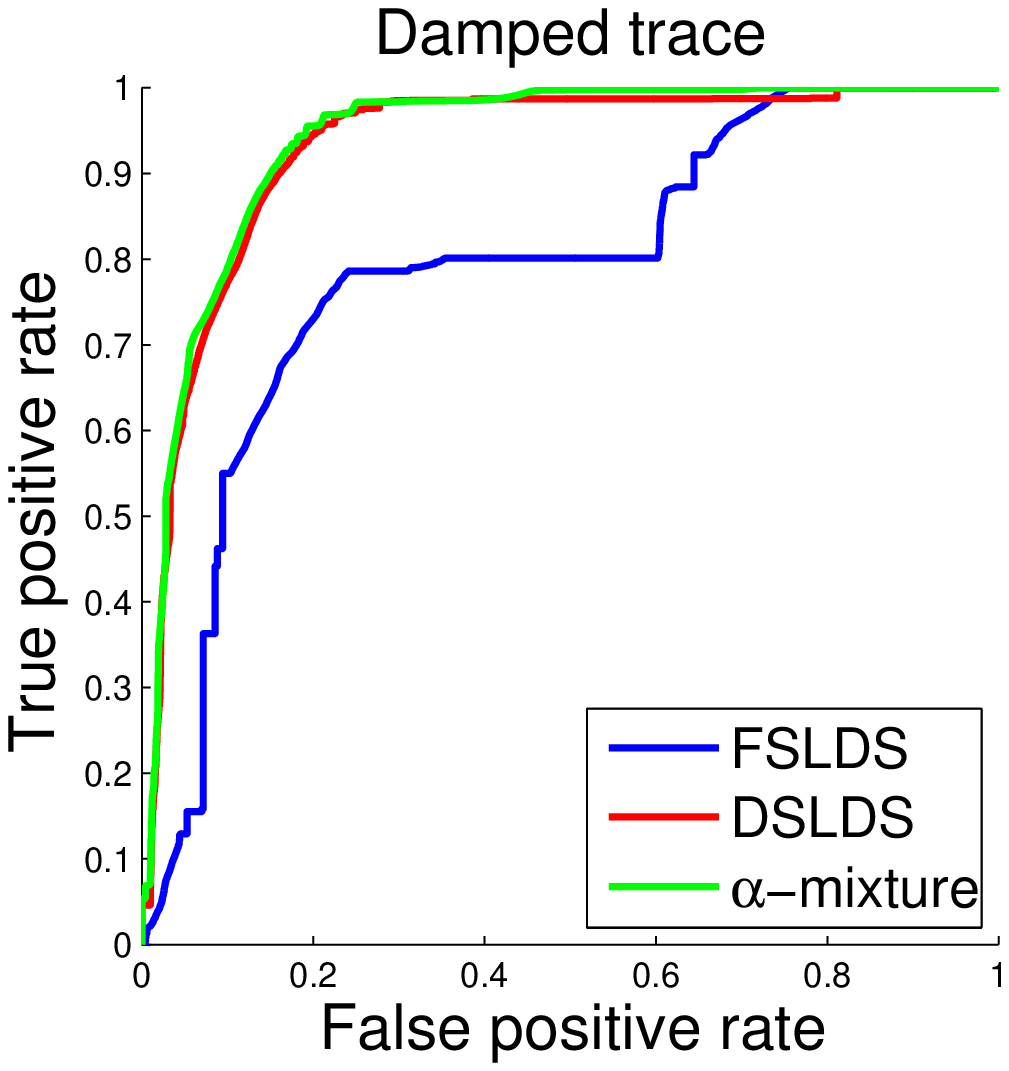}
        \end{subfigure}

        \begin{subfigure}[b]{0.25\textwidth}
                \centering
                \includegraphics[width=\textwidth, height = 6.5cm, keepaspectratio=true]{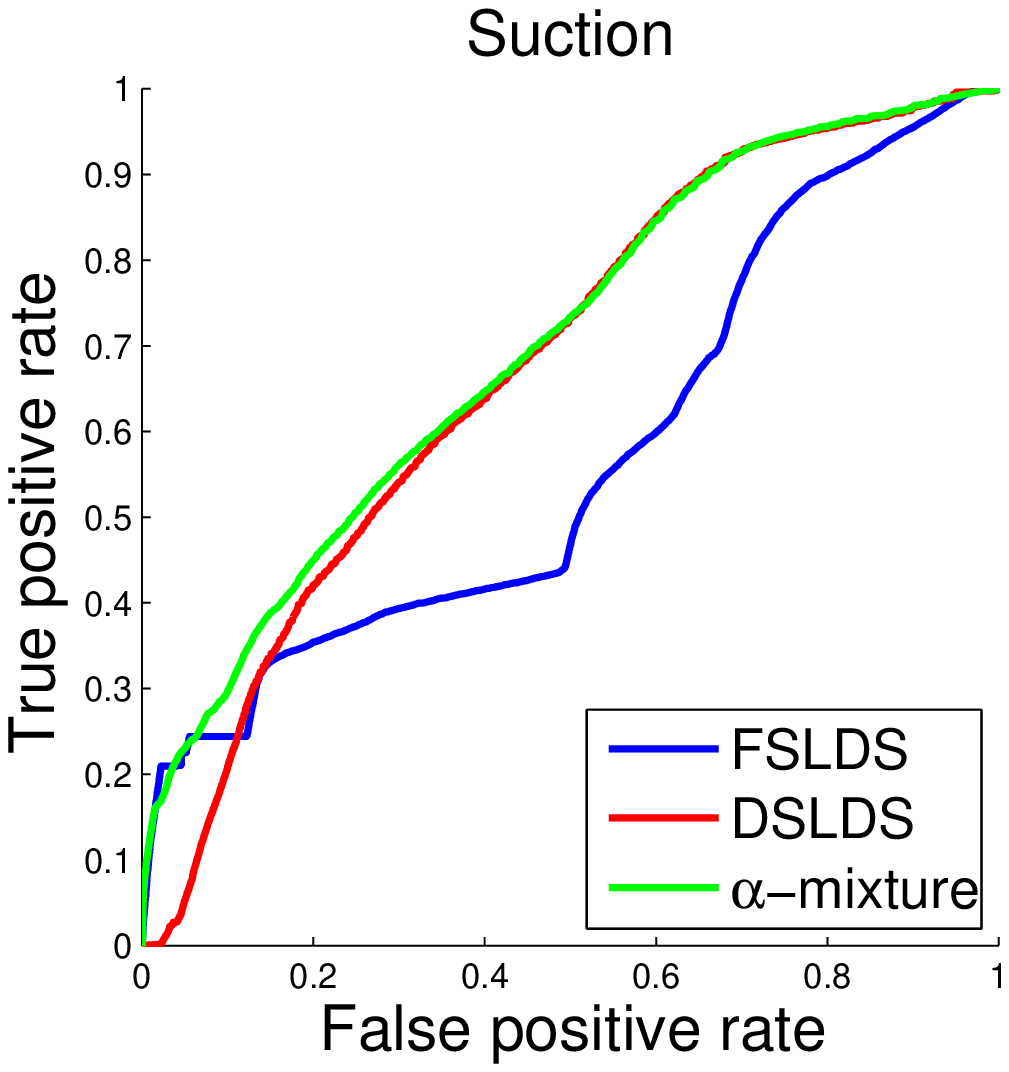}
        \end{subfigure}%
        ~      
        \begin{subfigure}[b]{0.25\textwidth}
                \centering
                \includegraphics[width=\textwidth, height = 6.5cm, keepaspectratio=true]{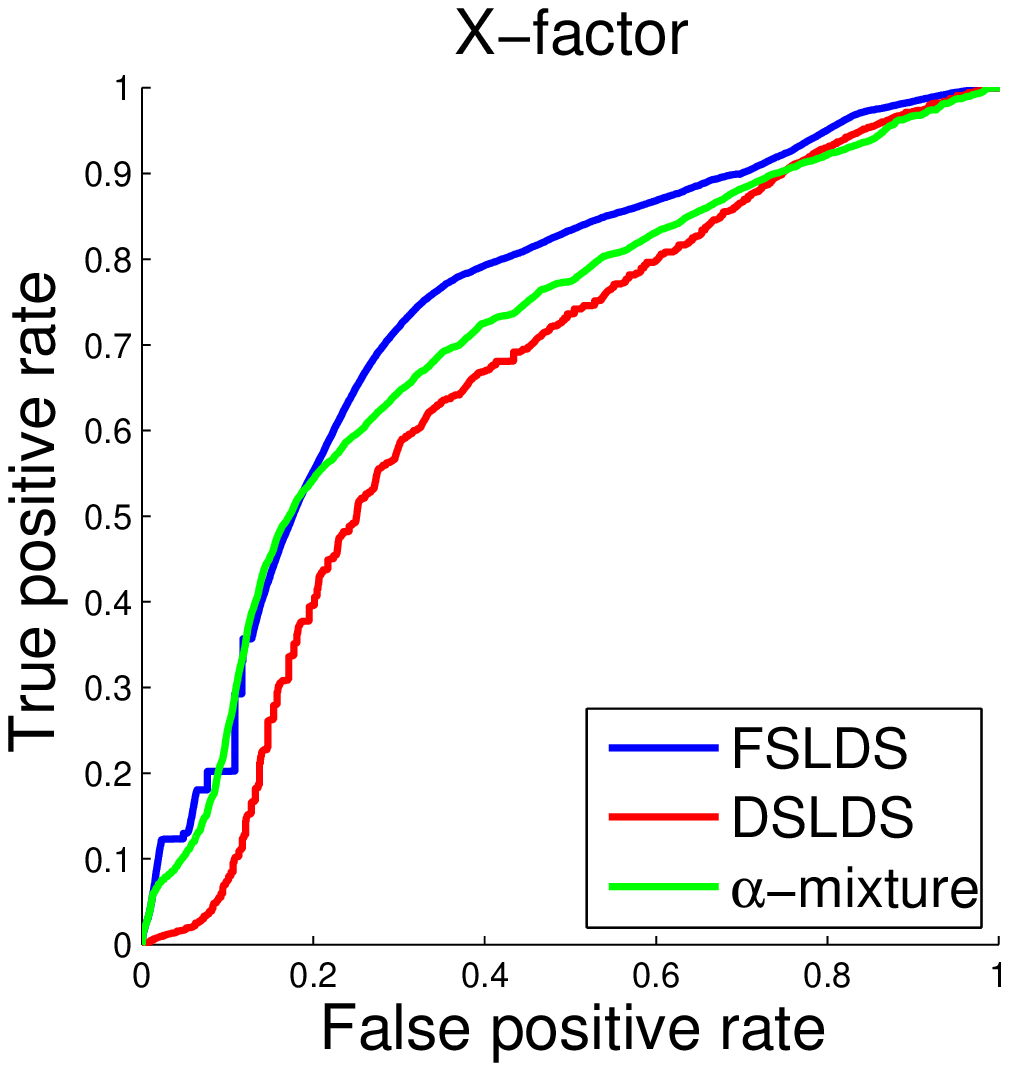}
        \end{subfigure}
        \caption{ROC curves per modelled factor in the case of the adult ICU.}
        \label{fig:ROC_sgh}
        \end{center}
\end{figure}

For both datasets we compare the performance of the DSLDS and the FSLDS for the task of inferring a patient's state of health. We measure the performance of the models by reporting the Area under the Receiver Operating Characteristic curve (AUC). Also, in Figures \ref{fig:ROC_nicu} and \ref{fig:ROC_sgh}, we provide plots of the Receiver Operating Characteristic curves (ROC) for the classification of the factors of interest comparing the DSLDS, the FSLDS, and an $\alpha$-mixture of the two models.

In the case of the DSLDS, the features described in Section \ref{sec:features} involve a number of hyperparameters that need to be chosen. Fitting them with a standard cross-validation (CV) scheme when data are not abundant poses a non-negligible risk of overfitting. As is shown in \citet{varma2006bias}, using CV to evaluate performance of a model when the model's hyperparameters have been themselves tuned using CV can lead to an optimistic bias of the estimate of the true performance. In that same work, a nested CV approach is shown to yield an almost unbiased estimate of the true performance, which we also follow in our experiments. In the outer loop the data are partitioned into $P$ disjoint test sets. After choosing one of these partitions, the rest of the data are used in the inner loop in a standard CV setup to select the hyperparameters. The hyperparameters which yielded the highest performance (average cross-validated AUC across factors in our case) in the inner loop are then used to estimate the performance of the model on the partition (test set) in the outer loop. This process is repeated $P$ times, once for each partition in the outer loop. For both datasets, we use leave-one-patient-out CV for the inner loop and 3-fold CV for the outer loop. In the inner loop, we perform a grid search over hyperparameters in the following sets: a) number of trees of random forest classifiers in \{10, 25, 50, 100, 200\}; b) $l$ in \{4, 9, 14, 19, 29, 49\}; c) $r$ in \{0, 5, 10\}. The sub-segments lengths (for slope features) were always set to max\{5, ($l+r+1$)/5\} and the kernel widths (for moving average features) were always set to max\{5, ($l+r+1$)/5\}.

In the case of the FSLDS, it is not necessary to follow the same procedure. Using the AIC score, as shown in Section \ref{sec:learning}, for choosing the orders of the ARIMA processes (which constitute the model's hyperparameters) avoids potential overfitting by penalising the model's likelihood as the parameters grow. We therefore use 3-fold CV to evaluate the FSLDS's performance.  

To evaluate the $\alpha$-mixture model, we have chosen the optimal $\alpha$ value as the one that maximises the average AUC across factors, via 3-fold CV. This also allowed us to explore the behaviour of the model as a function of $\alpha$ for both datasets. 

\subsubsection{Neonatal ICU}
\label{sec:NICU_res}

\begin{figure}[ht]
        \begin{center}
        \begin{subfigure}[b]{0.5\textwidth}
                \centering
                \hspace{0.2\textwidth} {Neonatal ICU}
                \includegraphics[width=\textwidth, height = 5cm]{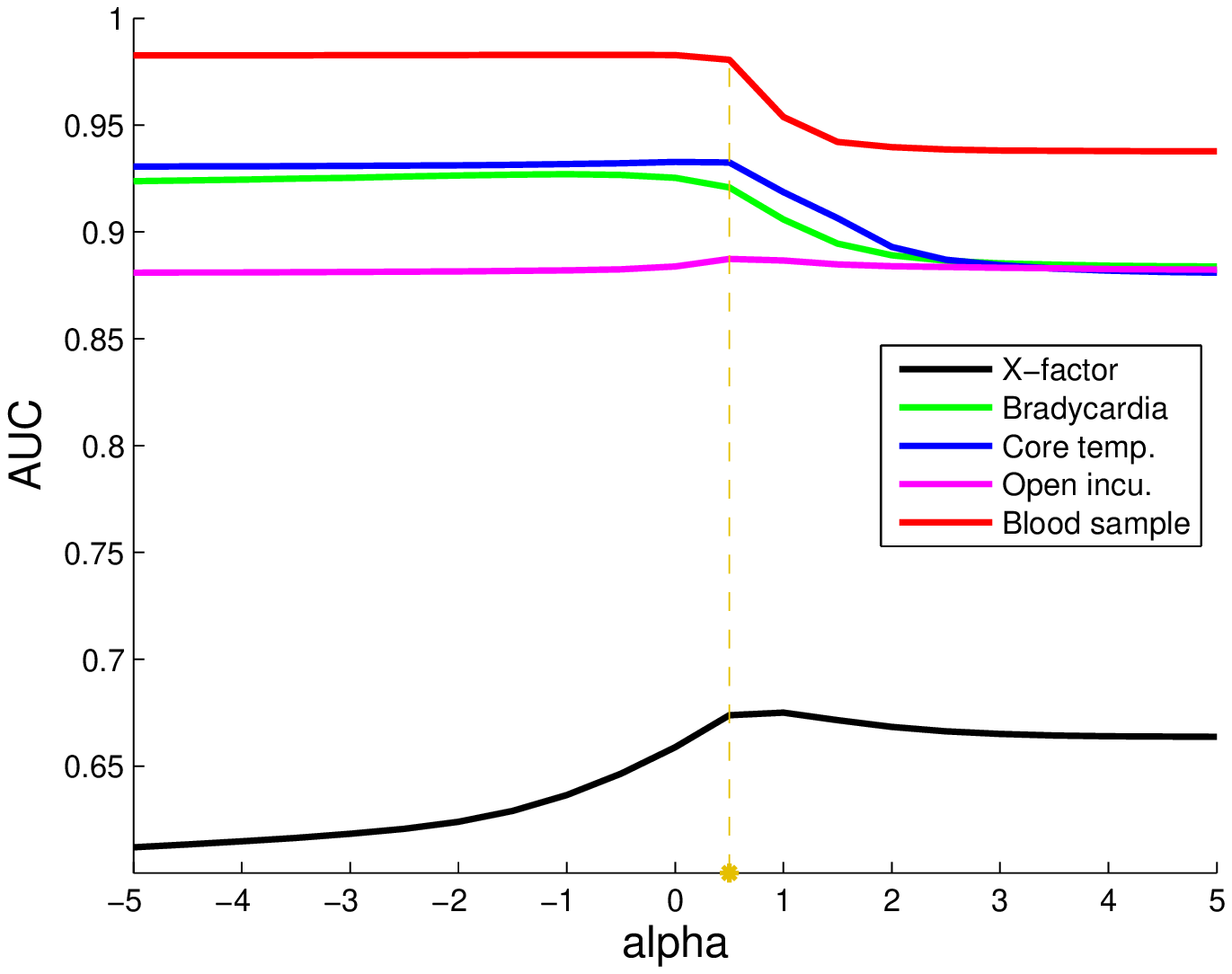}
        \end{subfigure}%
         
        \begin{subfigure}[b]{0.5\textwidth}
                \centering
                \hspace{0.2\textwidth} {Adult ICU}
                \includegraphics[width=\textwidth, height = 5cm]{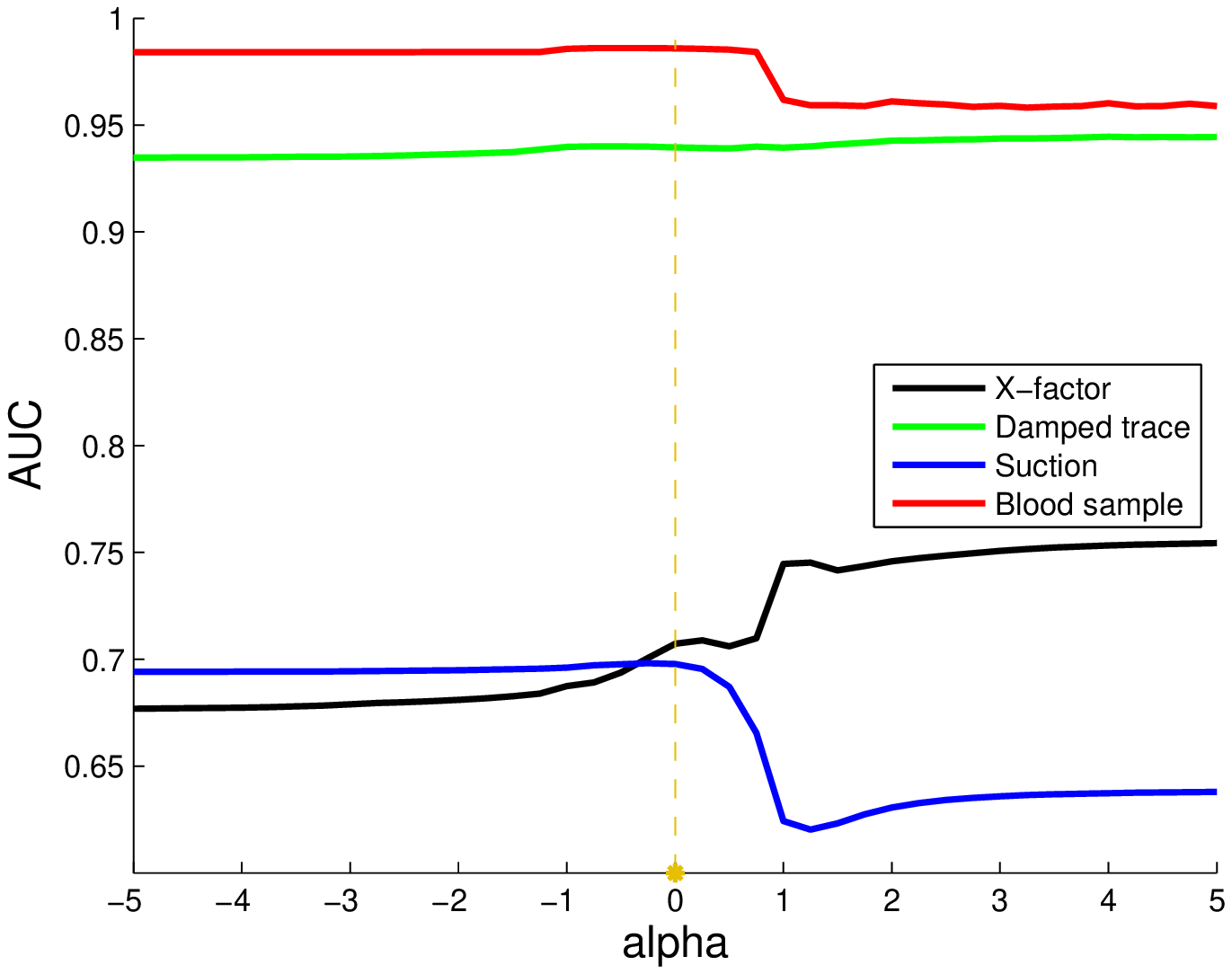}
        \end{subfigure}
        \caption{Performance of the $\alpha$-mixture models as a function of $\alpha$ ($step=0.25$) for the Adult ICU (top) and the neonatal ICU dataset (bottom). The asterisk marks the optimal value for $\alpha$.}
        \label{fig:alpha_vs_AUC}
        \end{center}
\end{figure}

In the case of the neonatal ICU we compare the two models on the full set of annotated factors reported in \citet{quinn2009factorial}. The results are shown in Table \ref{tab:NICU}\footnote{The FSLDS results were obtained using code provided by \citet{quinn2009factorial} with the same parameters as the ones mentioned there. The results are very close with the exception of the core temperature disconnection factor (for which the reported AUC in \citet{quinn2009factorial} was 0.79, while we obtained a value of 0.88), and the blood sample factor (for which the reported AUC in \citet{quinn2009factorial} was 0.96, while we obtained a value of 0.92).}. The DSLDS outperforms the FSLDS in three out of the four clinically identified factors. The difference in favour of the DSLDS is clear for bradycardias and blood samples, but less pronounced for core temperature disconnections. The FSLDS achieves slightly higher performance in the case of the incubator open factor, and clearly outperforms the DSLDS in the case of the X-factor. The FSLDS models the presence of outliers by the inclusion of an extra factor, which is essentially governed by the same parameters as stability with the only difference being that the system noise covariance is an inflated version of the respective covariance of the stability dynamics (for more details, see \citealp{quinn2009factorial}). Such an approach has the potential to address the issue of outlier detection in a more general and thus more satisfactory way. In the case of the DSLDS, our approach is to collectively treat all abnormal events, other than the ones attributed to known factors, as an ``X-class'' and build a binary classifier to distinguish that class. As the training datapoints for this class are highly inhomogeneous in terms of shared discriminative features, and test points belonging to the X-class may not exhibit a high degree of similarity to the training set, it is not surprising that the DSLDS may perform rather poorly for the X-factor. However, by considering an $\alpha$-mixture of the two models, we can combine the discriminative power of the DSLDS for known factors with the increased performance of the FSLDS for the X-factor, thus achieving a higher performance (bottom line of Table \ref{tab:NICU}) compared to considering the two models separately. The behaviour of the $\alpha$-mixture model as a function of $\alpha$ is shown in Figure \ref{fig:alpha_vs_AUC} (top). The optimal $\alpha$-mixture ($\alpha = 0.5$) yields the best average AUC across factors (in fact, $\alpha=0.5$ yields optimal performance for each factor separately except bradycardia, where it is almost optimal) compared to all other considered $\alpha$ values and also outperforms the DSLDS and the FSLDS in all cases except for the bradycardia factor, where the DSLDS performs slightly better. 

\begin{figure}[ht]
        \begin{center}
        \begin{subfigure}[b]{0.5\textwidth}
                \centering
                \includegraphics[width=\textwidth, height = 4cm]{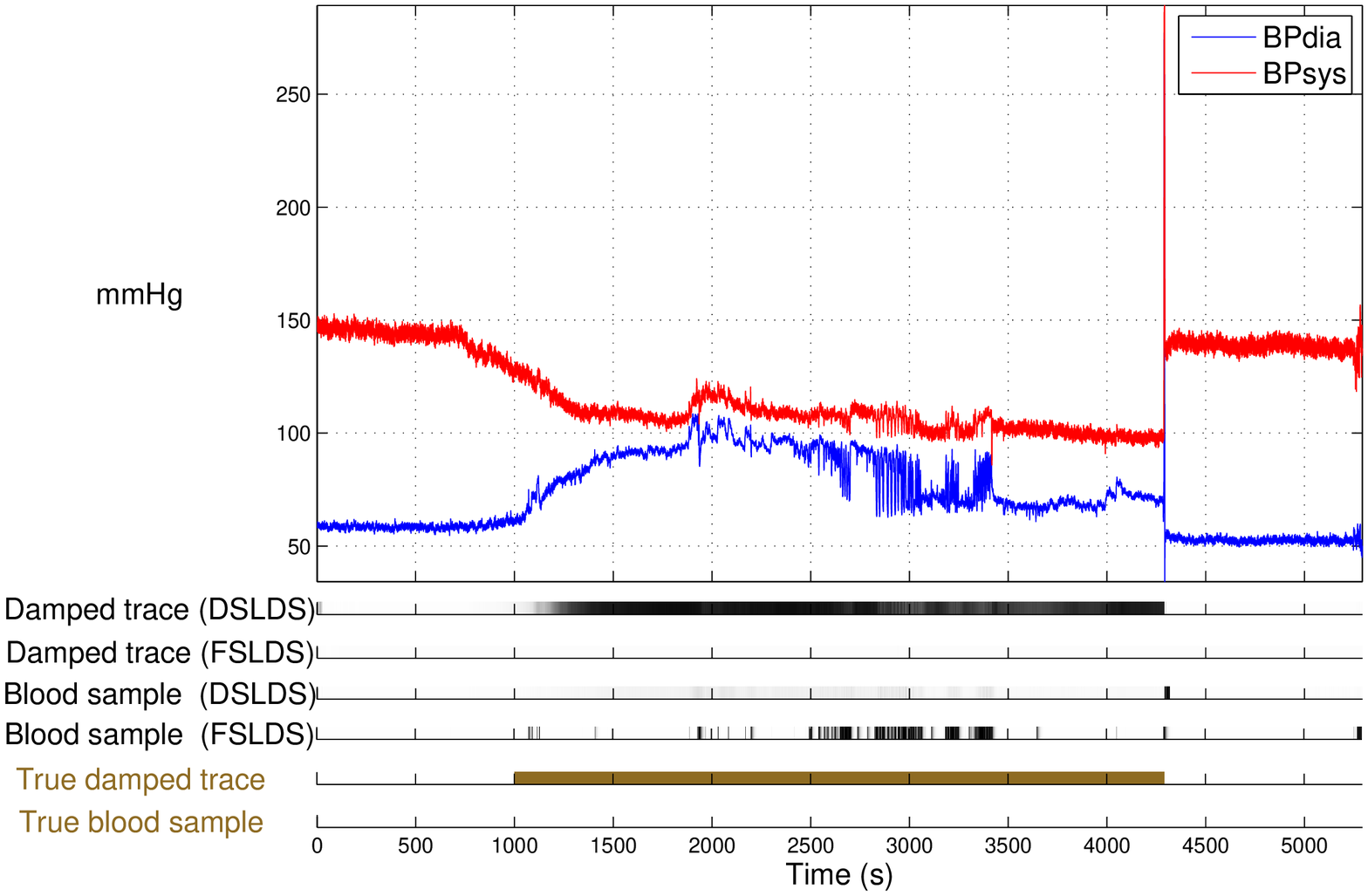}
        \end{subfigure}%
         
        \begin{subfigure}[b]{0.5\textwidth}
                \centering
                \includegraphics[width=\textwidth, height = 4cm]{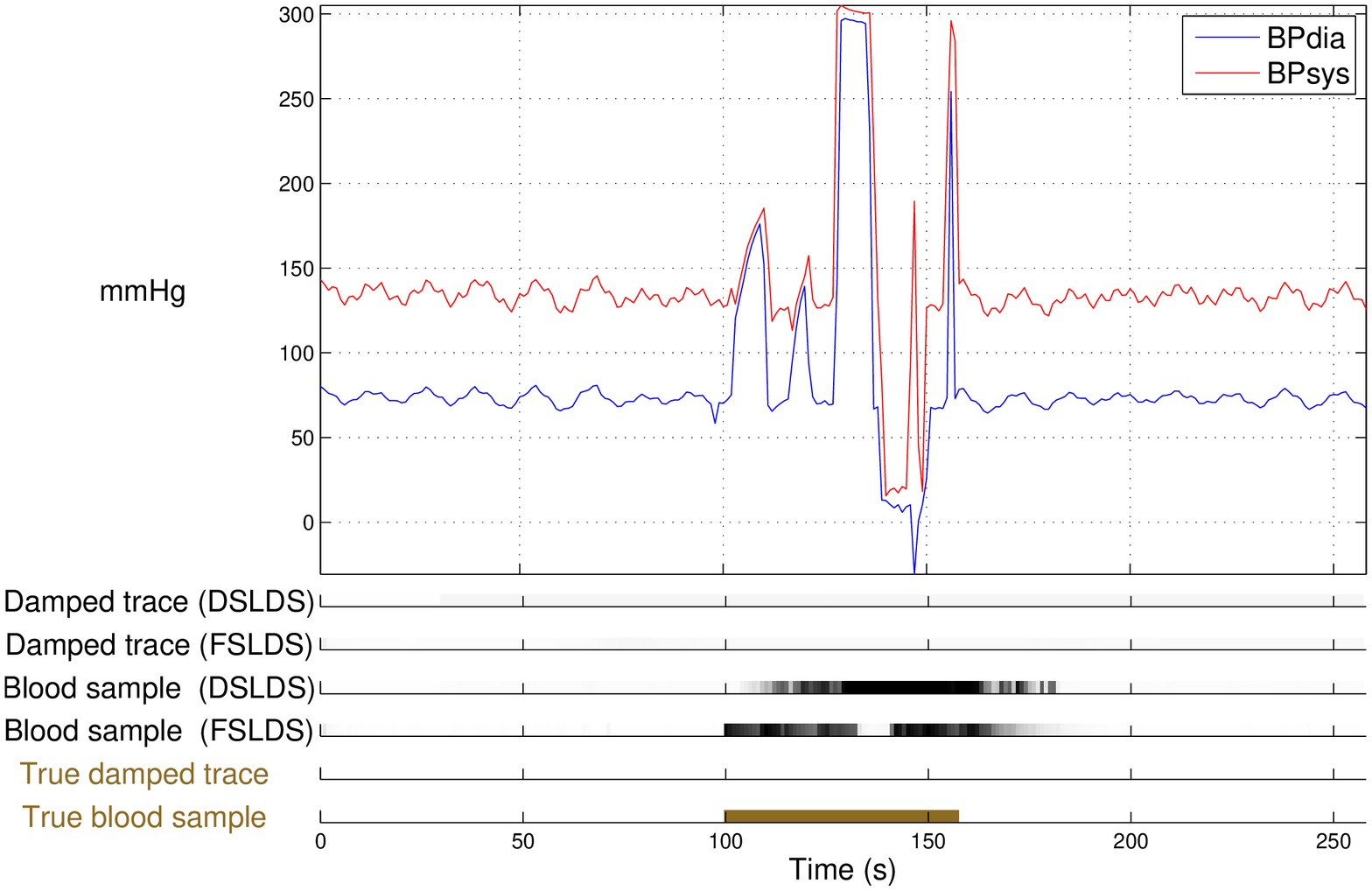}
        \end{subfigure}
        \caption{Example of DSLDS and FSLDS inferences for a damped trace event (top) and a blood sample event (bottom).}
        \label{fig:SGH_damped_BS}
        \end{center}
\end{figure}

\subsubsection{Adult ICU}
\label{sec:SGH_res}

\begin{table}
\caption{Comparison of DSLDS, FSLDS and $\alpha$-mixture performance for the Adult ICU dataset. Optimal value of the $\alpha$ parameter is shown inside parenthesis.}
\label{tab:SGH}
\begin{center}
\begin{tabular}{c l l l l}
\multicolumn{1}{c}{\bf AUC}   &\multicolumn{1}{c}{BS} &\multicolumn{1}{c}{DT} &\multicolumn{1}{c}{SC} &\multicolumn{1}{c}{X}
\\ \hline \\
DSLDS                      \hspace{2mm}    &$0.96$   \hspace{1mm}      &$0.93$ \hspace{1mm}        &$0.67$  \hspace{1mm}       &$0.65$\\      
FSLDS                      \hspace{2mm}    &$0.95$         &$0.79$         &$0.57$         &$0.74$\\
$\alpha$-mixture$^{(0)}$   \hspace{2mm}    &$0.99$         &$0.94$         &$0.70$         &$0.71$\\
\end{tabular}
\end{center}
\end{table}

In the case of the adult ICU, inferences for two example events are shown in Figure \ref{fig:SGH_damped_BS}. In the top, a damped trace event is shown, which lasts for almost one hour before being resolved by a flushing event (spiking of both channels). The DSLDS accurately identifies the damped trace event, while the FSLDS fails totally to detect it, but hypothesises several incorrect blood sample events. In the bottom panel a blood sample event is shown, where the multiple stages are clearly visible. The event starts with two artifactual ramps, followed by a flushing, a zeroing, and finally with another flushing. This is slightly different than the description we have already given, but slight deviations from the standard protocol due to human error is to be expected. In this case, both models manage to capture the event in a generally satisfactory manner. Summary results are reported in Table \ref{tab:SGH}. The DSLDS outperforms the FLSDS on all of the known factors. The damped trace and suction events particularly are characterised by high variability which is hard to capture with a generative process. However, simple discriminative features are able to capture them with higher accuracy. As was expected, the FSLDS achieves a higher AUC for the X-factor. Again, the optimal $\alpha$-mixture ($\alpha = 0$) outperforms the DSLDS and the FSLDS in all cases except for the X-factor, where the FSLDS achieves a slightly higher AUC. Contrary to the neonatal ICU dataset, as shown in Figure \ref{fig:alpha_vs_AUC} (bottom) there are alternative $\alpha$ values which can yield higher AUC across different factors. For example, an X-factor AUC value of 0.76 can be obtained by setting $\alpha=5$. However, apart from the superior (on average) performance of the $\alpha$-mixture, another appealing property is that $\alpha$ could be treated as a user-tunable parameter. In a practical setting, the model could be preset with the optimal $\alpha$ value, but a clinician could decide, for example, to make the model focus on maximising its predictive performance on the X-factor (or some important physiological factor like bradycardia) to the potential detriment of other factors. Then the model could adjust its $\alpha$ parameter in real-time based on training data results to maximise its performance on the desired factor. 

\subsubsection{Inference for $\vc{x}$-state}
\label{sec:x_state_results}

\begin{figure}[ht]
        \begin{center}
        \hspace{-5mm}
        \begin{subfigure}[b]{0.5\textwidth}
                \centering
                \hspace{0.2\textwidth} {DSLDS}
                \includegraphics[width=\textwidth, height = 5cm]{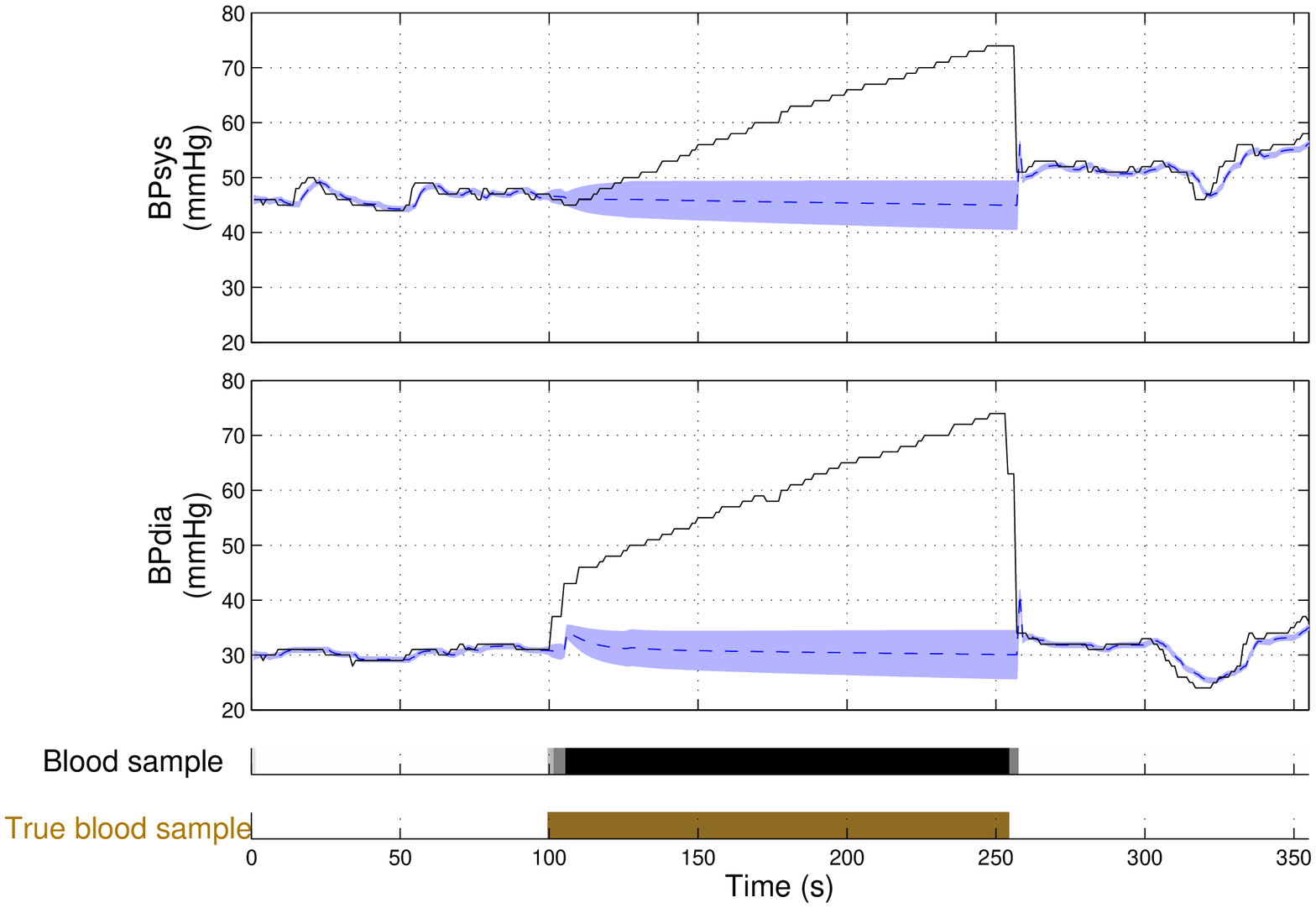}
        \end{subfigure}%

        \hspace{-5mm}      
        \begin{subfigure}[b]{0.5\textwidth}
                \centering
                \hspace{0.2\textwidth} {FSLDS}
                \includegraphics[width=\textwidth, height = 5cm]{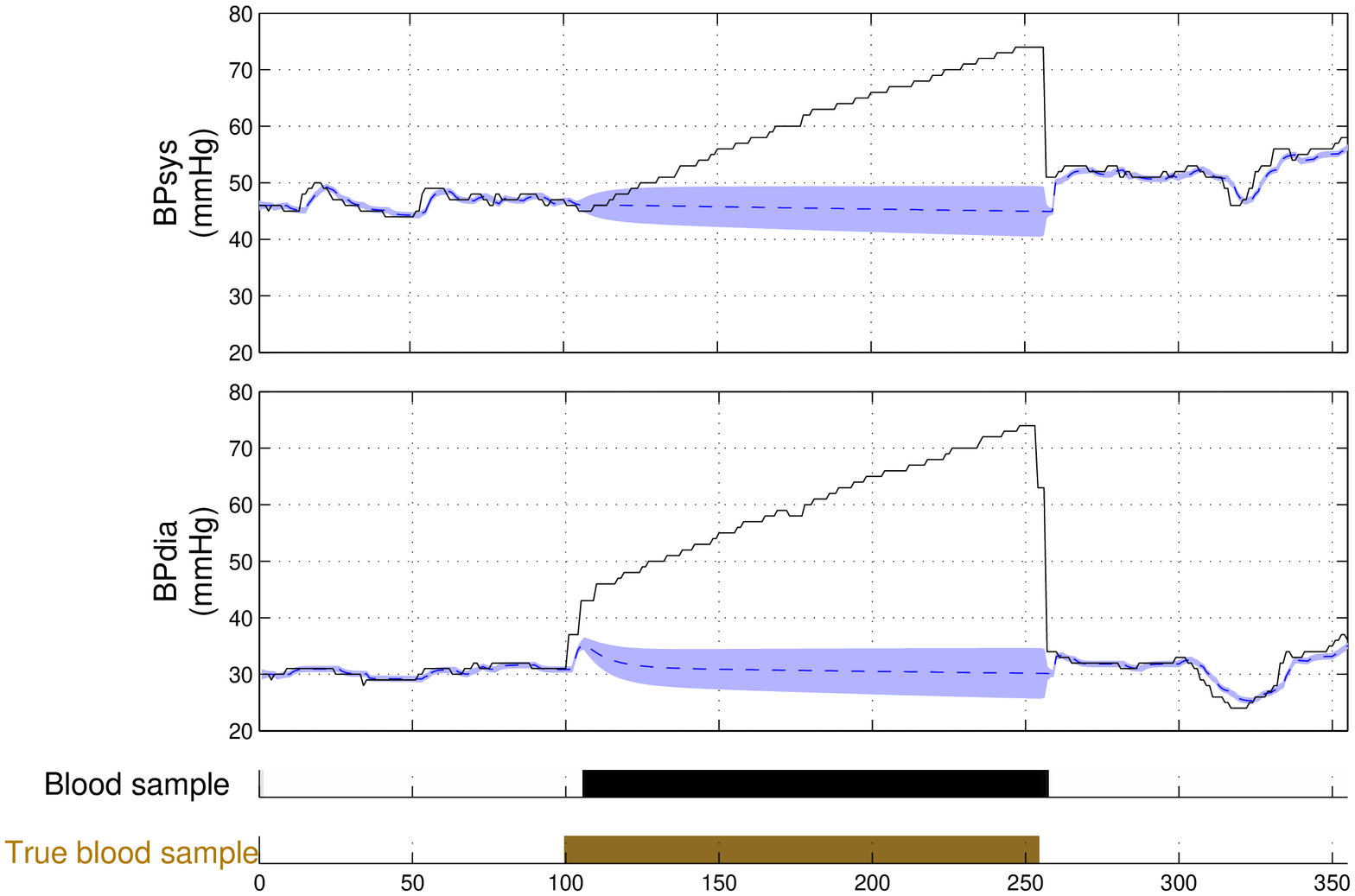}
        \end{subfigure}
        \caption{Example of the inferred underlying physiology in the presence of a blood sample in the case of the DSLDS (top) and the FSLDS (bottom). The solid line corresponds to the actual observations, while the estimated true physiology is plotted as a dashed line with the shaded area indicating two standard deviations.}
        \label{fig:x_state}
        \end{center}
\end{figure}

Finally, Figure \ref{fig:x_state} shows the inferred distribution of underlying physiology during a blood sample taken from a neonate for both models. In both cases, estimates are propagated with increased uncertainty under the correctly inferred artifactual event. Note a small difference at the start of the event: The DSLDS partially identifies the event causing an increase in uncertainty, while the FSLDS (incorrectly) identifies this part as stable and thus its $\vc{x}$-state update exhibits lower uncertainty. Maintaining an estimate of the underlying vital signs in the presence of artifacts can then be used for data imputation. Another use, which has been deemed important by our clinical experts, is that such an estimate can help doctors maintain an approximate view of a patient's underlying physiology during artifactual events that would otherwise completely obscure a patient's vital signs. This can be crucial during treatment of a patient under critical conditions, such as the ones found in an ICU. 

\section{Discussion}
\label{sec:discussion}
We have presented a discriminative approach for the very important application of patient monitoring in ICUs. We show that our new approach is able to outperform the previous generative approach used for the same task in most of the investigated cases. We also show that an $\alpha$-mixture of the two approaches yields better results than either model separately. In our approach we have assumed that the prediction of the switching variable factorises over the state space. However, one could use a structured output model to predict the joint distribution of different factors.

Finally, another issue is the lack of explicit temporal continuity in the $s$-chain. Implicitly, this is handled by the feature construction process. However, a future direction could be to establish a Markovian connection on the $s$-chain too and compare with our current approach. 

\subsubsection*{Acknowledgements}
We extend our thanks to Ian Piper and Christopher Hawthorne for their expert insight and annotated data, and to Martin Shaw and Partha Lal for preprocessing code and valuable discussions. Author KG was funded by the Scottish Informatics and Computer Science Alliance. This research was funded in part by the Chief Scientist Office (Scotland) ref.\ CZH/4/801.

\subsubsection*{References}
\bibliographystyle{natbib}
\renewcommand{\section}[2]{}
\bibliography{dslds.bib}

\begin{thebibliography}{}

\bibitem[Akaike(1972)Akaike]{akaike1972information}
Akaike, H. (1972).
\newblock Information theory and an extension of the maximum likelihood
  principle.
\newblock {\em 2nd Int. Symp. Information Theory, Supp. to Problems of Control
  and Information Theory\/}, pages 267--281.

\bibitem[Alspach and Sorenson(1972)Alspach and Sorenson]{alspach1972nonlinear}
Alspach, D.~L. and Sorenson, H.~W. (1972).
\newblock {Nonlinear Bayesian Estimation Using Gaussian Sum Approximations}.
\newblock {\em Automatic Control, IEEE Transactions on\/}, {\bf 17}(4),
  439--448.

\bibitem[Amari(2007)Amari]{amari2007integration}
Amari, S.-i. (2007).
\newblock {Integration of Stochastic Models by Minimizing $\alpha$-Divergence}.
\newblock {\em Neural Computation\/}, {\bf 19}(10), 2780--2796.

\bibitem[Breiman(2001)Breiman]{breiman2001random}
Breiman, L. (2001).
\newblock {Random Forests}.
\newblock {\em Machine Learning\/}, {\bf 45}(1), 5--32.

\bibitem[Brockwell and Davis(2009)Brockwell and Davis]{brockwell2009time}
Brockwell, P.~J. and Davis, R.~A. (2009).
\newblock {\em Time Series: Theory and Methods\/}.
\newblock Springer.

\bibitem[Diggle(1990)Diggle]{diggle1990time}
Diggle, P. (1990).
\newblock {\em Time Series: A Biostatistical Introduction\/}.
\newblock Oxford University Press.

\bibitem[Ghahramani and Hinton(1996)Ghahramani and
  Hinton]{ghahramani1996parameter}
Ghahramani, Z. and Hinton, G.~E. (1996).
\newblock {Parameter Estimation for Linear Dynamical Systems}.
\newblock Technical report, Technical Report CRG-TR-96-2, University of
  Totronto, Dept. of Computer Science.

\bibitem[Goldberger {\em et~al.}(2000)Goldberger, Amaral, Glass, Hausdorff,
  Ivanov, Mark, Mietus, Moody, Peng, and Stanley]{goldberger2000physiobank}
Goldberger, A.~L., Amaral, L.~A., Glass, L., Hausdorff, J.~M., Ivanov, P.~C.,
  Mark, R.~G., Mietus, J.~E., Moody, G.~B., Peng, C.-K., and Stanley, H.~E.
  (2000).
\newblock Physiobank, physiotoolkit, and physionet components of a new research
  resource for complex physiologic signals.
\newblock {\em Circulation\/}, {\bf 101}(23), e215--e220.

\bibitem[Hastie {\em et~al.}(2009)Hastie, Tibshirani, and
  Friedman]{hastie2009elements}
Hastie, T., Tibshirani, R., and Friedman, J. (2009).
\newblock {\em {The Elements of Statistical Learning}\/}.
\newblock Springer.

\bibitem[Lafferty {\em et~al.}(2001)Lafferty, McCallum, and
  Pereira]{lafferty2001conditional}
Lafferty, J., McCallum, A., and Pereira, F.~C. (2001).
\newblock Conditional random fields: Probabilistic models for segmenting and
  labeling sequence data.
\newblock In {\em International Conference on Machine Learning (ICML)\/}.

\bibitem[Lehman {\em et~al.}(2014)Lehman, Adams, Mayaud, Moody, Malhotra, Mark,
  and Nemati]{lehman2014physiological}
Lehman, L., Adams, R., Mayaud, L., Moody, G., Malhotra, A., Mark, R., and
  Nemati, S. (2014).
\newblock A physiological time series dynamics-based approach topatient
  monitoring and outcome prediction.
\newblock {\em Biomedical and Health Informatics\/}.

\bibitem[Lerner and Parr(2001)Lerner and Parr]{lerner2001inference}
Lerner, U. and Parr, R. (2001).
\newblock {Inference in Hybrid Networks: Theoretical Limits and Practical
  Algorithms}.
\newblock In {\em {Proceedings of the Seventeenth conference on Uncertainty in
  Artificial Intelligence (UAI)}\/}, pages 310--318. Morgan Kaufmann Publishers
  Inc.

\bibitem[Lu {\em et~al.}(2009)Lu, Murphy, Little, Sheffer, and
  Fu]{lu2009hybrid}
Lu, W.-L., Murphy, K.~P., Little, J.~J., Sheffer, A., and Fu, H. (2009).
\newblock {A Hybrid Conditional Random Field for Estimating the Underlying
  Ground Surface from Airborne LiDAR Data}.
\newblock {\em Geoscience and Remote Sensing, IEEE Transactions on\/}, {\bf
  47}(8), 2913--2922.

\bibitem[McCallum {\em et~al.}(2000)McCallum, Freitag, and
  Pereira]{mccallum2000maximum}
McCallum, A., Freitag, D., and Pereira, F.~C. (2000).
\newblock {Maximum Entropy Markov Models for Information Extraction and
  Segmentation.}
\newblock In {\em {International Conference on Machine Learning (ICML)}\/},
  pages 591--598.

\bibitem[Murphy(1998)Murphy]{murphy1998switching}
Murphy, K.~P. (1998).
\newblock {Switching Kalman Filters}.
\newblock Technical report, U.C.Berkeley.

\bibitem[Quinn {\em et~al.}(2009)Quinn, Williams, and
  McIntosh]{quinn2009factorial}
Quinn, J.~A., Williams, C.~K., and McIntosh, N. (2009).
\newblock {Factorial Switching Linear Dynamical Systems applied to
  Physiological Condition Monitoring}.
\newblock {\em Pattern Analysis and Machine Intelligence, IEEE Transactions
  on\/}, {\bf 31}(9), 1537--1551.

\bibitem[S{\"a}rkk{\"a}(2013)S{\"a}rkk{\"a}]{sarkka2013bayesian}
S{\"a}rkk{\"a}, S. (2013).
\newblock {\em Bayesian Filtering and Smoothing\/}.
\newblock Cambridge University Press.

\bibitem[Shumway and Stoffer(2000)Shumway and Stoffer]{shumway2000time}
Shumway, R.~H. and Stoffer, D.~S. (2000).
\newblock {\em {Time Series Analysis and Its Applications}\/}.
\newblock Springer New York.

\bibitem[Varma and Simon(2006)Varma and Simon]{varma2006bias}
Varma, S. and Simon, R. (2006).
\newblock Bias in error estimation when using cross-validation for model
  selection.
\newblock {\em BMC Bioinformatics\/}, {\bf 7}(1), 91.

\bibitem[Williams and Stanculescu(2011)Williams and
  Stanculescu]{williams2011automating}
Williams, C.~K. and Stanculescu, I. (2011).
\newblock {Automating the Calibration of a Neonatal Condition Monitoring
  System}.
\newblock In {\em {Artificial Intelligence in Medicine}\/}, pages 240--249.
  Springer.

\end{thebibliography}

\end{document}